\documentclass{article}

\usepackage[main,final]{neurips_2026}

\usepackage[utf8]{inputenc} 
\usepackage[T1]{fontenc}    
\usepackage{hyperref}       
\usepackage{url}            
\usepackage{booktabs}       
\usepackage{amsfonts}       
\usepackage{nicefrac}       
\usepackage{microtype}      
\usepackage{xcolor}         
\usepackage{amsmath}
\usepackage{graphicx} 
\usepackage{subcaption}

\usepackage[capitalize,noabbrev]{cleveref}

\usepackage[textsize=tiny]{todonotes}

\usepackage{multirow}
\usepackage{csquotes}
\crefname{equation}{Eq.}{Eqs.}
\usepackage{pifont}
\newcommand{\cmark}{\ding{51}} 
\newcommand{\xmark}{\ding{55}}
\usepackage{wrapfig}
\usepackage{algorithm}
\usepackage{algorithmic}

\title{CHARM+: Cross-Hardware Attention with Re-Merge Multistream Mechanism}

\newcommand{\blankauthor}{\phantom{Xinhang Zhang}}
\author{%
  Xinhang Zhang\thanks{Equal contribution.}
  \And Boning Zhang\footnotemark[1]
  \And Chengchun Liu
  \And Chunpu Li
  \AND
  Lei Sun
  \And Weifeng Zhang
  \And Limin Xiao\thanks{Corresponding author.}
  \And \blankauthor
  \\[-1em]
  \makebox[\dimexpr\textwidth-2\tabcolsep\relax][c]{Lenovo Research}
}

\begin{document}

\maketitle

\begin{abstract}
  As sequence lengths continue to grow, distributed attention based on sequence parallelism (e.g., Ring-Attention) has become a mainstream solution for large language models (LLMs). Nevertheless, inherent communication dependencies limit computation–communication overlap, reducing GPU utilization. Moreover, in cross-hardware (NUMA or node) architectures, due to imbalanced communication bandwidth, frequent communication synchronization will limit overall performance. Based on these insights, we propose CHARM+, Cross-Hardware Attention with Re-Merge Multistream Mechanism. We reorganize sequential communications into several levels and achieve intra-level parallelism and computation overlap. Furthermore, we decouple cross-hardware communication from intra-hardware communication, eliminating synchronization overhead caused by bandwidth imbalance. Forward and backward algorithms were evaluated on multiple hardware architectures and consistently outperformed state-of-the-art methods. Compared to Ring-Attention, our forward and backward algorithms achieve average speedups of $1.3\times$ and $1.7\times$ on a dual-NUMA PCIe architecture, and $2.8\times $ and $3.9\times$ on a dual-node NVLink architecture, respectively.
\end{abstract}

\section{Introduction}
\label{sec:1}

As large language models (LLMs) scale up \cite{gpt3}, longer context lengths are required to fully exploit their performance \cite{LLMSurvey, LLMSurvey2}. Notably, LLM training requires processing long documents to enhance model capabilities \cite{longcontextTraining}. Meanwhile, the inference of LLMs involves cross-document understanding \cite{long-context-understanding}, long program comprehension \cite{long-code-comprehension}, and multi-turn interactive dialogues \cite{multi-turn-conversation}. To handle this, self-attention is the core mechanism for capturing long-range dependencies using queries ($q$), keys ($k$), and values ($v$) \cite{longRangeDependencies,transformer}. However, its computation and memory costs scale with sequence length, making long-context training and inference expensive \cite{longContextChallenge,Duo-attention-inference}. Therefore, sequence parallelism (SP) and distributed attention have become mainstream approaches. They partition the sequence across accelerators to reduce per-device overhead, enabling longer context windows \cite{distributedAttention,sequenceParallel}.

Nevertheless, existing SP methods (e.g., Ring-Attention) exhibit two limitations. 
Firstly, due to the communication dependency, each communication relies on the completion of the preceding
communication \cite{communicationDependency}. Consequently, it limits the communication-computation overlap and reduces GPU utilization.
Secondly, in the cross-hardware (Non-Uniform Memory Access, NUMA or node), imbalanced bandwidth leads to extra synchronization overhead \cite{imbalancedBandwidth}. Therefore, communication dependencies cause GPUs to idle, degrading overall performance.

\begin{wrapfigure}{l}{0.55\textwidth} 
    \centering
    \includegraphics[width=0.53\textwidth]{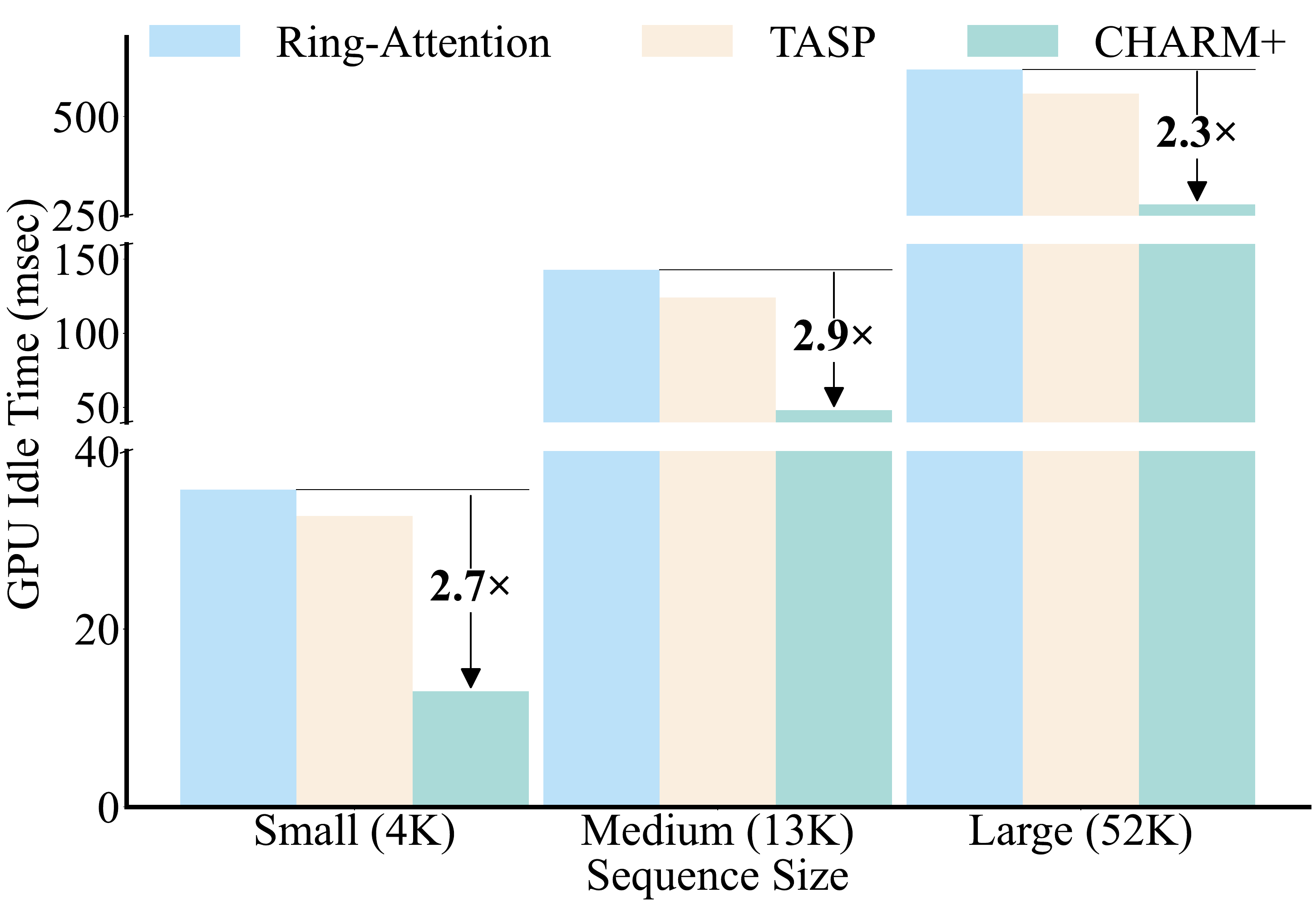} 
    \caption{GPU Idle Time. By setting various sequence lengths, we compare GPU idle time with Ring-Attention and TASP. Our method achieves higher GPU utilization.}
    \label{fig:GPU_idle_time}
    \vspace{-0.8cm}
\end{wrapfigure}

Based on these observations, we propose CHARM+, Cross-Hardware Attention with Re-Merge Multistream Mechanism. We design a merge-based communication mechanism to reorganize the 
sequential communications into several levels and achieve intra-level independence. Secondly, we leverage multiple communication streams to better overlap computation and communication. Furthermore, by decoupling intra- and inter-hardware communication, we avoid performance degradation due to synchronization overhead. As shown in \cref{fig:GPU_idle_time}, GPU idle time has been significantly reduced, leading to substantial performance improvements. Our contributions are summarized as follows:

\begin{itemize}
\item \textbf{We propose a novel communication and parallel mechanism.}
To optimize the ring pattern, we reorganize communications into logarithmic levels and design a multistream parallelism for $kv$ and $q$ to achieve intra-level parallelism.
\item \textbf{We develop a Cross-Hardware Query Parallelism to decouple intra- and inter-hardware communication.} By isolating cross-hardware $q$ and $kv$ communications, the parallelism avoids synchronization induced by bandwidth imbalance and improves performance.
\item \textbf{We implement both forward and backward of our method and conduct extensive experiments across different hardware architectures.} The results demonstrate that our forward and backward consistently outperform Ring-Attention, achieving maximum speedups of $1.4\times$ and $1.9\times$ on PCIe, and $3.2\times$ and $4.3\times$ on NVLink.

\end{itemize}

\section{Related Work}
\label{sec:2}

\subsection{Interconnection Architecture}

\begin{figure}[t]
    \centering
    \begin{subfigure}[b]{0.49\columnwidth}
        \centering
        \includegraphics[width=\columnwidth]{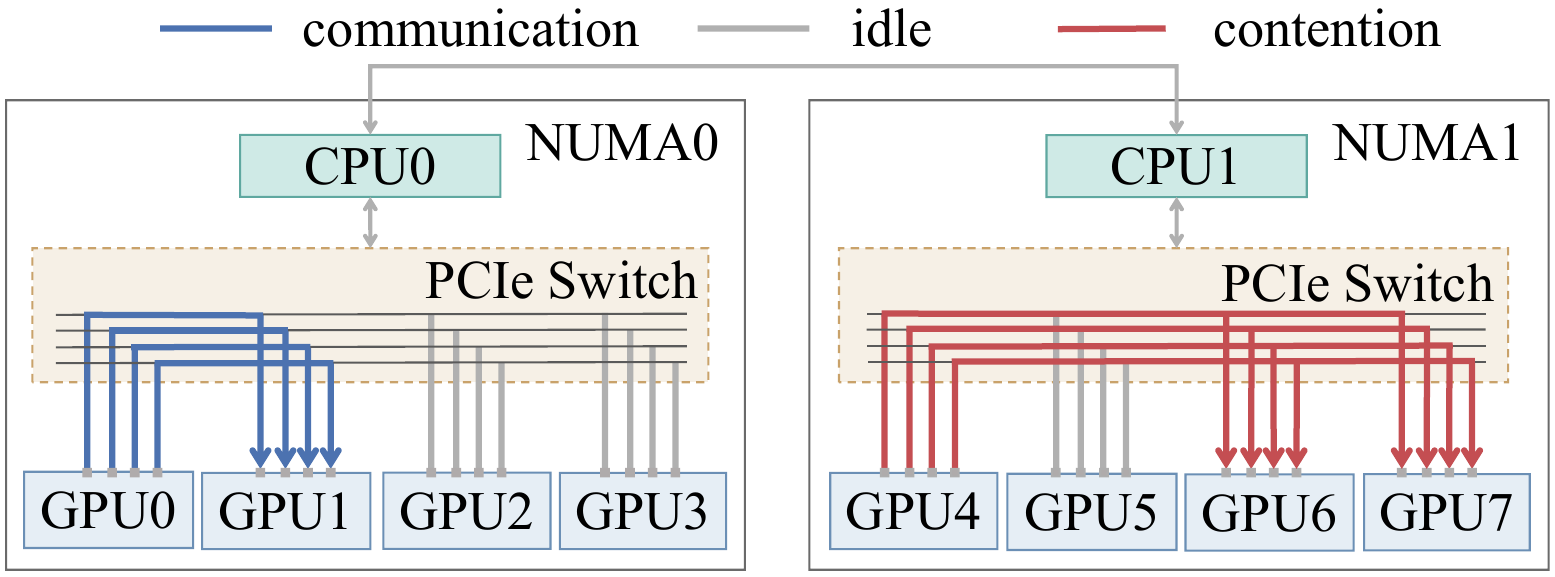}
        \caption{Blocking comm due to link contention in PCIe}
        \label{fig:pcie}
    \end{subfigure}
    \hfill
    \begin{subfigure}[b]{0.49\columnwidth}
        \centering
        \includegraphics[width=\columnwidth]{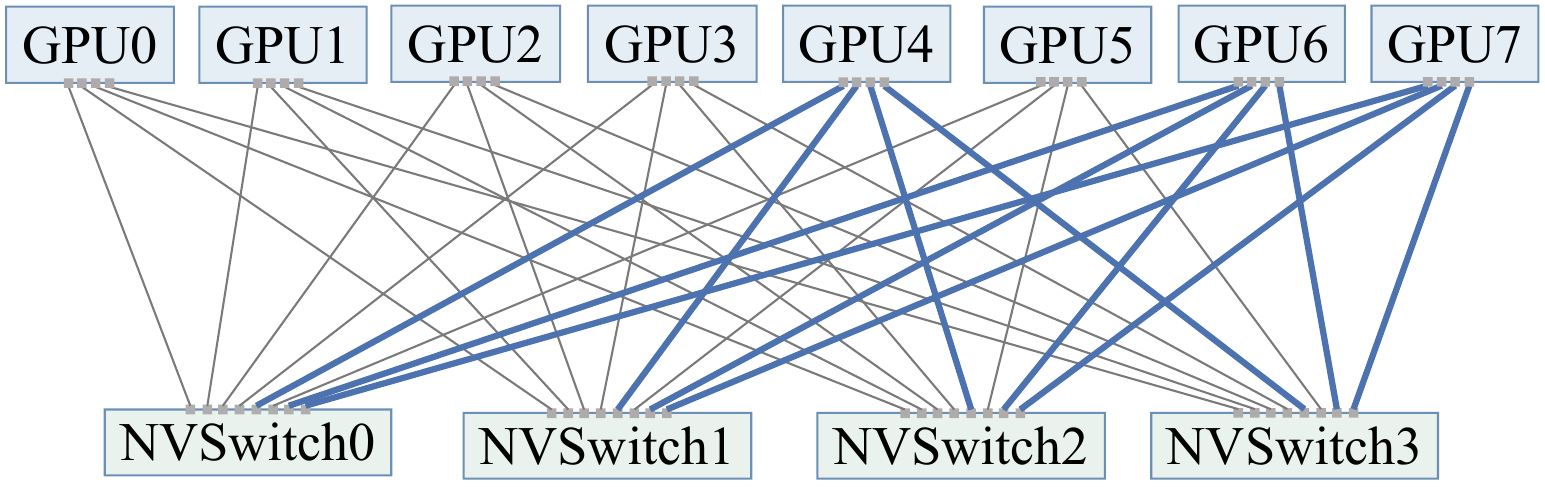}
        \caption{Non-blocking comm on dedicated links in NVLink}
        \label{fig:nvlink}
    \end{subfigure}

    \caption{Communication under PCIe and NVLink. The left of (a) shows single communication at full bandwidth without contention, while the right shows multiple communications at limited bandwidth with contention. (b) shows multiple communications at full bandwidth without contention.}
    \vspace{-0.5cm}
    \label{fig:different_interconnect}
\end{figure}

Most large-scale training and inference are deployed on multi-GPU systems connected via either PCIe or NVLink \cite{introducePcie,introduceNvlink}. \cref{fig:different_interconnect} illustrates the interconnects of these two architectures. PCIe-based 
systems often suffer from link contention due to concurrent communication requests, while 
NVLink provides high-bandwidth, dedicated links that enable contention-free communication \cite{pcie-contention-analysis}. Furthermore, both architectures have imbalanced communication bandwidth between cross-hardware (i.e., cross-NUMA or cross-node)  and intra-hardware \cite{crossandIntraHardware}. Consequently, it incurs substantial synchronization overhead, severely degrading communication performance.


 \subsection{Sequence Parallelism}
\begin{figure*}[t]
  \vskip 0.2in
  \begin{center}
    \centerline{\includegraphics[width=\textwidth]{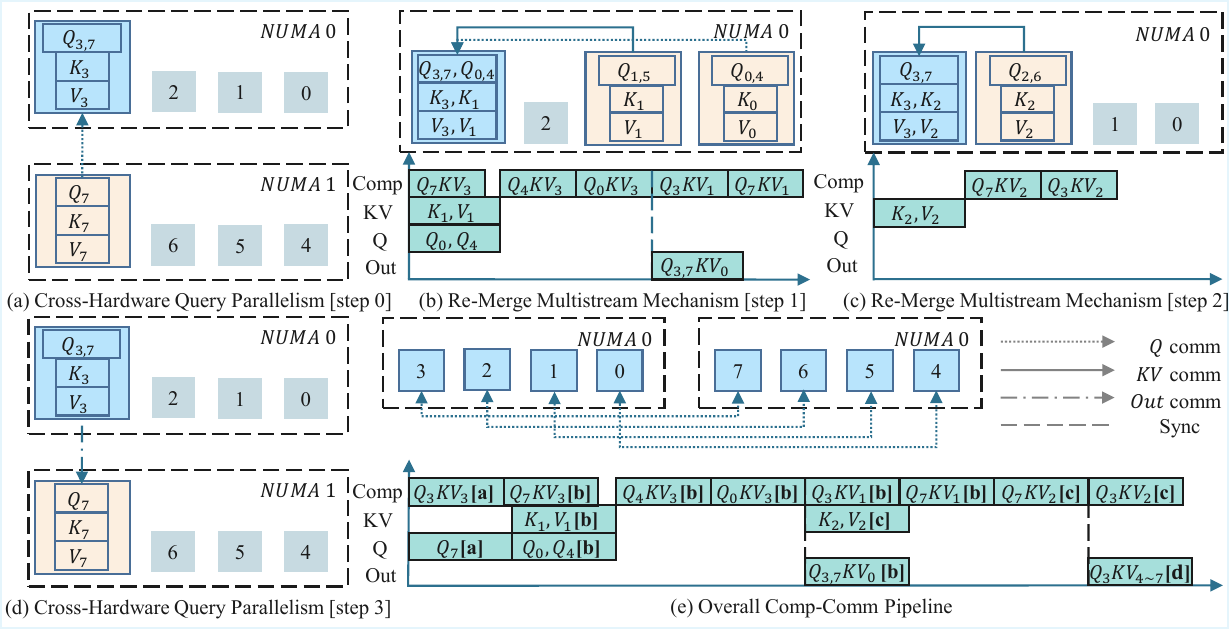}}
    \caption{
      Overall Framework of CHARM+. (a) - (d) illustrate the complete process of the algorithm using an 8-GPU system with a dual-NUMA configuration.  
      (a) displays parallel communication of $q$ is first performed across NUMA. As a result, GPU 3 receives the $q$ from GPU 7, eliminating the need for further cross-NUMA communication. 
      (b) shows the parallel communication and computation of $q$ and $kv$. In this step, GPU 3 communicates $q$ with GPU 0 and $kv$ with GPU 1. After completing the computation of $Q_{0,4}$, GPU3 communicates back to GPU 0 and receives the output.
      (c) performs the last step of  RM+, where GPU 3 only communicates $kv$ with GPU 2.  
      (d) shows that once all $kv$ results are computed, the output is communicated back to GPU 7.   
      (e) illustrates the overall cross-NUMA communication and the full overlap of computation and communication of GPU3 happened in (a)-(d). 
    }
    \vspace{-0.7cm}
    \label{fig:overall_framework}
  \end{center}
\end{figure*}
As long-context requirements in LLMs continue to grow, the sequence parallelism and the distributed attention mechanism have become critical techniques for scaling attention by distributing sequence computation across multiple devices \cite{sequenceParallel,distributedAttention}. Existing methods (e.g., Ring-Attention) rely on distributed $kv$ exchange, where $kv$ blocks are transmitted across devices in a ring topology \cite{ringTopology} and processed incrementally \cite{ring-attention}. However, there exists a dependency in ring-based communication: each communication cannot start until the previous one completes. Moreover, each communication can only overlap with a single computation \cite{commCompOverlap}. Consequently, as communication time grows, this dependency prolongs GPU idle time and reduces effective compute utilization.

Several approaches have been proposed to improve the efficiency and scalability of Ring-Attention. TASP \cite{TASP} aims to better utilize available communication links by adopting the all-to-all communication. Attention-2D \cite{attention-2d} splits the self-attention computation into two dimensions along $q$ and $kv$. Mesh-Attention \cite{mesh-attention} replaces one-dimensional Ring-Attention with a two-dimensional tiled assignment. Striped-Attention \cite{striped-attention} reorders the input sequence into uniformly distributed blocks. However, these approaches do not optimize the ring-based pattern. Reducing the ring length or enabling multiple rings to communicate in parallel does not eliminate the communication dependencies, nor does it improve the communication–computation overlap. Therefore, efficient computation–communication overlap and high GPU utilization remain challenges in distributed attention.

\section{Method}
\label{sec:3}





 

This section presents our method illustrated in \cref{fig:overall_framework}\footnote{For clarity, the computation of the FlashAttention in the illustration will ignore causal masks.}. We first explain our motivation and introduce a merge-based communication mechanism. Secondly, we propose a multistream parallelism to improve the communication-computation pattern in the communication mechanism. Furthermore, we leverage a cross-hardware query parallelism to mitigate the imbalanced bandwidth between cross-hardware (NUMA or node) communication and intra-hardware communication and use intra-pipeline to hide cross-hardware communication. All notations utilized in this section are described in \cref{tab:notation_description}.
\subsection{Motivation}
\label{sec:3_1}


    As introduced in \cref{sec:2}, the current ring-based attention involves gradually passing $kv$ through a ring path for attention computation and update. This mechanism enables partial overlap between communication and computation. However, dependencies between communications prevent a communication from starting before the previous one is complete. To maximize GPU utilization, we seek to maximize communication–computation overlap, defined as follows: 
    \begin{align}
        \min{O} &= \sum_{i=1}^{N} \ell{(T_{\text{comm}}^{i}(k,v) - T_{\text{comp}}^{i}(q,k,v))}
    \end{align}
    Let $N$ denote the length of critical path that is non-overlapping, $T_{\text{comm}}^{i}(\cdot)$ and $T_{\text{comp}}^{i}(\cdot)$ are the $i_{\text{th}}$ communication time, the $i_{\text{th}}$ computation time of FlashAttention \cite{flashattention1}. We define $O$ as the unhidden communication time, where the GPU is idle. Therefore, minimizing $O$ maximizes GPU utilization.

    \begin{table}[t]
    \centering
    \caption{Notation Description.}
    \label{tab:notation_description}
    \begin{tabular}{c|c||c|c}
    \toprule
    Notation & Description & Notation & Description \\
    \midrule
    $Q, q$ & Query matrix 
    & $K, k$ & Key matrix \\
    
    $V, v$ & Value matrix 
    & $n$ & Degree of sequence parallelism \\
    
    $O$ & Unhidden communication time 
    & $\ell(\cdot)$ & Function that clips value to zero \\
    
    $T_{\text{comm}}(\cdot)$ & Communication time of a tensor 
    & $T_{\text{comp}}(\cdot)$ & Computation time of Attention \\
    \bottomrule
    \end{tabular}
    \vspace{-0.4cm}
    \end{table}

    For Ring-Attention \cite{ring-attention}, due to the ring structure, this objective can be expressed as:
    \begin{align}
    \label{eq:ring_overlap}
        O_{\text{ring}} &= \sum_{i=1}^{n-1} \ell(T_{\text{comm}}^{i}(k, v) - T_{\text{comp}}^{i}(q, k, v))
    \end{align}
    This $O(n)$ critical path dependency becomes the primary bottleneck limiting the ability to overlap communication and computation. In particular, the bandwidth imbalance between cross-hardware (NUMA or node) and intra-hardware communication causes extra communication synchronization to significantly increase the latency of each step along the critical path. Therefore, this effect further exacerbates the overall bottleneck, we further corroborate these observations with Nsight profiling results in \cref{fig:nsight_ring} of \cref{sec:nsight_compare}. Based on these observations, our motivation becomes:
    \begin{displayquote}
    ``\textit{How can we efficiently overlap computation and communication while simultaneously minimizing the overall critical path of the execution?} "
    \end{displayquote}
    
\begin{figure}[t]
    \centering
    \begin{subfigure}[b]{0.49\textwidth}
        \centering
        \includegraphics[width=\columnwidth]{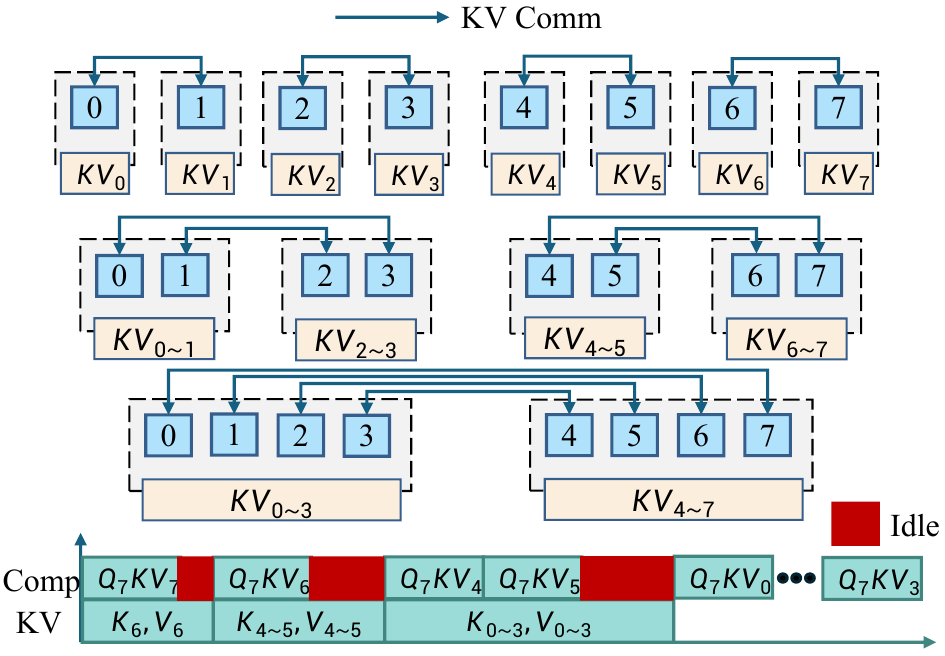}
        \caption{Merge Communication Mechanism.}
        \label{fig:merge}
    \end{subfigure}
    \hfill
    \begin{subfigure}[b]{0.49\textwidth}
        \centering
        \includegraphics[width=\columnwidth]{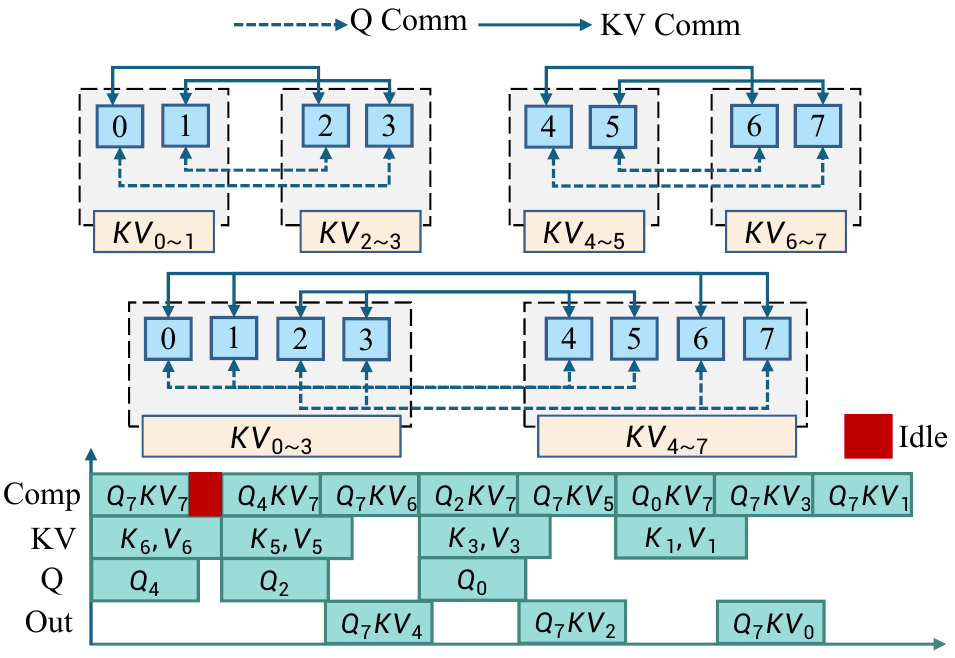}
        \caption{Re-Merge Multistream Mechanism.}
        \label{fig:remerge}
    \end{subfigure}

    \caption{The Computation and Communication Overlap on GPU 7. For 8 GPUs, (a) contains three levels. Each GPU communicates at groups of 1, 2, and 4 at different levels, aggregating the communicated $kv$ while overlapping computation. (b) maintains consistency with (a) in the first levels, but further enables multistream communication and computation of $q$ and $kv$ at other each merge level. Each GPU concurrently communicates $q$ and $kv$ with different GPUs, computes the $q$ of other GPU and communicates the output back. The red region indicates GPU idle time.}
    \vspace{-0.4cm}
    \label{fig:merge_framework}
\end{figure}

    As inspired by the traditional merge algorithm \cite{mergetree,traditionalMerge}, we present a communication mechanism. Compared to the ring-based $kv$ exchange, the mechanism redesigns this process as $kv$ aggregation. The aggregation is structured as a logarithmic-level merge (in \cref{fig:merge}, a sequence parallelism of 8 corresponds to a three-level merge). At each level $i$, $kv$ are aggregated in parallel between device pairs with a stride of $2^{i-1}$. As communication dependencies persist between merge levels, the critical path of the merge mechanism becomes $O(\log n)$, which is significantly shorter than Ring-Attention.
    
    For the merge communication mechanism, we model the communication time at each merge level as:
    \begin{align}
    \label{eq:merge_comm}
        C^{(1)}_{i} &= \sum_{j=1}^{2^{i-1}} T_{\text{comm}}^{j}(k, v)
    \end{align}
    At the $i_{\text{th}}$ merge level, the communication mechanism performs $2^{i-1}$ communications. Meanwhile, the computation at level $i$ processes the $kv$ received at the communication of level $i-1$. Thus, KV communication at level $i$ overlaps with computation at level $i-1$, whose cost is $2^{\ell(i-2)}$. Consequently, the unhidden communication time is given by:
    
    \begin{align}
    \label{eq:merge_overlap}
        O_{\text{merge}} &= \sum_{i=1}^{\log_2(n)} \ell(C^{(1)}_{i} - \sum_{j=1}^{2^{\ell({i-2})}} T_{\text{comp}}^{j}(q, k, v))
    \end{align}
    As can be observed, compared to the Ring-Attention, the merge communication mechanism restructures the linear communication path into a merge path with logarithmic level, significantly shortening the communication critical path.  Each merge level accommodates more computation, increasing opportunities to overlap communication with computation and improving efficiency.
    
\subsection{Re-Merge Multistream Mechanism}
\label{sec:3_2}
The merge communication mechanism reduces dependencies between ring communications, but communication at each merge level still follows the ring pattern. To further increase communication–computation overlap, we design the Re-Merge Multistream Mechanism, named $\text{RM+}$. It still follows the structure of the merge communication mechanism, but optimizes the communication at each merge level to enable more communication-computation overlap.

As described in \cref{sec:3_1}, Ring-Attention involves communication only for $kv$, and each round of $kv$ communication can start only after the previous round completes. Notably, the communication of $q$ is independent of the $kv$ communication. By exploiting this independence, we parallelize $q$ communication with $kv$ communication, improving communication-computation overlap, as illustrated in \cref{fig:merge_framework}. Moreover, we aim to overlap the computation after the final merge level(e.g., $Q_{7}KV_{0}\cdot\cdot\cdot Q_{7}KV_{3}$ in \cref{fig:merge}. During each merge level, 
$kv$ is fetched sequentially from peer GPUs for computation. In addition, parallel communication of 
$q$ and $kv$ further reduces the communication time of each merge level. Consequently, as illustrated in \cref{fig:remerge}, we obtain:
\begin{align}
\label{eq:RM+_comm}
C^{(2)}_{i} &= \sum_{j=1}^{2^{\ell(i-2)}} 
        \max\left( T_{\text{comm}}^{j}(q),T_{\text{comm}}^{j}(k,v) \right)
\end{align}
Note that, \cref{eq:RM+_comm} follows \cref{eq:merge_comm}. Due to the parallel communication between $q$ and $kv$, the communication time is bounded by the maximum. Meanwhile, advancing the computation after the final merge level increases computations at earlier merge levels, thereby reducing the unhidden communication time. 
Specifically, all communication except the final one at each level can be overlapped, with the last communication deferred to the next merge level. Therefore, for the $i_{\text{th}}$ merge level, there are $2^{i-1}$ computations. 
Consequently, the unhidden communication time is:
\begin{align}
\label{eq:RM_overlap}
O_{\text{RM+}}
= \sum_{i=1}^{\log n }\ell(C^{(2)}_{i}-\sum_{j=1}^{2^{i-1}} T_{\text{comp}}^{j}(q,k,v))
\end{align}
Under the assumption that all parallel communications sustain identical bandwidth, \cref{eq:merge_overlap} outperforms \cref{eq:ring_overlap} due to its shorter critical path and greater communication overlap. However, parallel communication of multistream can lead to PCIe link contention, increasing latency. Consequently, on the PCIe architecture, the single communication of $\text{RM+}$ can be longer than that in Ring-Attention. Therefore, performance on a GPU system using NVLink is significantly better than on a PCIe architecture, as further demonstrated in \cref{sec:ablation_study}.

\subsection{Cross-Hardware Query Parallelism}
\label{sec:3_3}

In multi-hardware architectures (NUMA or Node), each KV communication in Ring-Attention simultaneously involves both intra-hardware and cross-hardware communication. Since cross-hardware\begin{wrapfigure}{l}{0.55\textwidth} 
    \centering
    \includegraphics[width=0.53\textwidth]{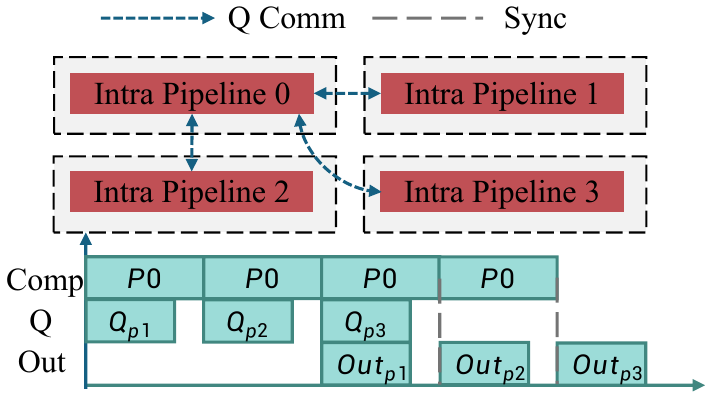} 
    \caption{Cross-Hardware Query Parallelism. The complete process is illustrated with pipeline 0 as an example. Each intra-pipeline processes queries (Q) from other pipelines and uses the same workflow with its local KV, returning the final out back.  Effective overlap allows inter-pipeline communication to be hidden by intra-pipeline, further improving performance.}
    \label{fig:Cross_Hardware_Parallel}
    \vspace{-0.3cm}
\end{wrapfigure}
communication typically provides lower bandwidth, the faster intra-hardware communication is forced to be idle while waiting for the communication synchronization, leading to substantial performance inefficiency.
As illustrated in \cref{fig:Cross_Hardware_Parallel}, we propose a novel parallelism where cross-hardware communication of the $q$ is interleaved with the intra-node pipeline. The pipelines process the current queries (Q) from different pipelines with local KV using the same computation and communication flow, as detailed in \cref{sec:3_2}.
 Notably, since all $kv$ and $q$ communications occur in the pipeline come from the same NUMA or node, the bottleneck caused by synchronized communication is largely eliminated. With effective overlap illustrated in \cref{fig:Cross_Hardware_Parallel}, after all computations are completed, only a final out communication is not hidden. Therefore, as the number of nodes increases, effective overlap ensures that only a single communication remains exposed, making the advantage of our method increasingly pronounced, which is further validated in the ablation study in \cref{sec:ablation_study}
 and \cref{sec:H100}. We provide Nsight profiling results of CHARM+ in \cref{fig:nsight_charm+} of \cref{sec:nsight_compare}, demonstrating that our overlap strategy significantly outperforms Ring-Attention.

\section{Experiments}
\label{sec:4}
\subsection{Experimental Setup}

We conduct experiments on GPU systems with different configurations and detailed specifications are provided in \cref{tab:main_experiment_setup}. We use official NCCL and RDMA test benchmarks to measure the intra-hardware
\begin{wraptable}{r}{0.55\textwidth}
  \centering
  \vspace{-0.2cm}
  \caption{The GPU Configuration and Bandwidth.}
  \label{tab:main_experiment_setup}
  \resizebox{0.55\columnwidth}{!}{
  \begin{tabular}{cccccc}
    \toprule
    \multirow{2}{*}{GPU} 
    & \multirow{2}{*}{NVLink} 
    & \multirow{2}{*}{Num} 
    & \multirow{2}{*}{TFLOPS} 
    & \multicolumn{2}{c}{Bandwidth (GB/s)} \\
    \cmidrule(lr){5-6}
      &   &   &   & Cross & Intra \\
    \midrule
    L20 & \xmark & 8  & 119.5 & 14.7 & 24.0  \\
    H20 & \checkmark & 16 & 148.0 & 42.6 & 297.2 \\
    H100 & \checkmark & 32 & 989.0 & 42.5 & 365.1 \\
    \bottomrule
  \end{tabular}%
  }
  \vspace{-0.5cm}
\end{wraptable}
and cross-hardware unidirectional P2P bandwidth. On L20, we evaluate the performance in both cross-NUMA and intra-NUMA settings. Furthermore, we explore the impact of link contention in PCIe-based systems. On H20 and H100, we evaluate cross-node and intra-node performance and further compare results under NVLink architecture, where link contention is absent.

We evaluate our method under diverse configurations by varying batch size ($B$), sequence length ($S$), and hidden dimensions ($H$). 
Specifically, sequence lengths range from 1K to 100K (up to 200K in some scenarios), with four hidden-dimension configurations formed by combining dimensions of 12 or 20 with sizes of 128 or 192. We conduct a grid search over $B$, $S$, and $H$, ensuring that the product $BS$ does not exceed 500K due to memory constraints. Each configuration is evaluated over 100 iterations following 20 warm-up iterations, with results averaged across the measured runs. To enable a unified characterization of different configurations, we use the computation-to-communication ratio (CCR) as a single metric. Following TASP \cite{TASP}, we define the CCR as:
\begin{equation}
\label{eq:ccr}
CCR=\sum_{i=1}^{N}T^{i}_{\text{comp}(q,k,v)}/\sum_{i=1}^{N}T^{i}_{\text{comm}(k,v)}
\end{equation}
Here, $q$, $k$, and $v$ are tensors that depend on the sequence length, the batch size, the hidden dimensions, and the number of attention heads. We accumulate the computation and communication time of Ring-Attention under sequential execution. The final metric is defined as the ratio of the computation time to the communication time (detailed analysis of CCR is provided in \cref{sec:CCR}). We compare our method against Ring-Attention \cite{ring-attention}, TASP, and their corresponding zigzag variants \cite{striped-attention}\footnote{As TASP does not provide the implementation of the backward algorithm, we compare forward performance only.}. 
\begin{figure*}[t]
    \centering
    \begin{minipage}{0.32\textwidth}
        \centering
        {\textbf{Forward}\par} 
        \includegraphics[width=\linewidth]{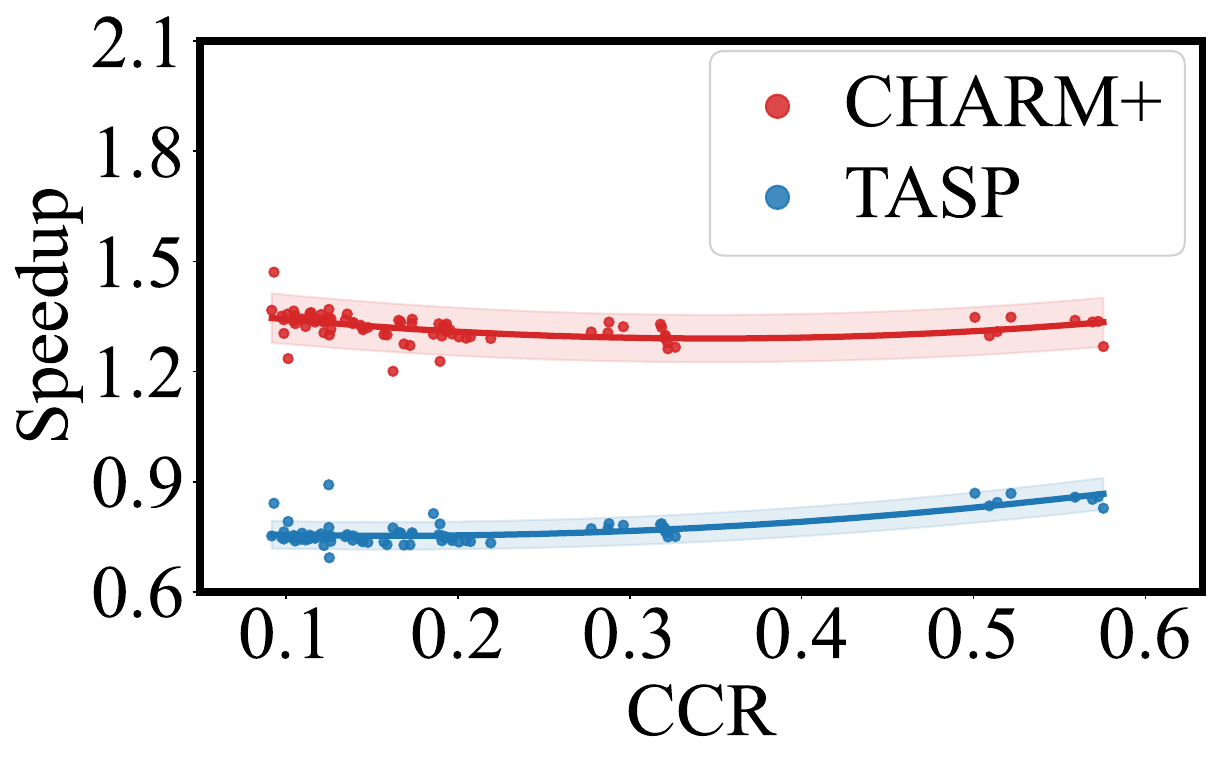}
    \end{minipage}\hfill
    \begin{minipage}{0.32\textwidth}
        \centering
        {\textbf{Backward}\par}
        \includegraphics[width=\linewidth]{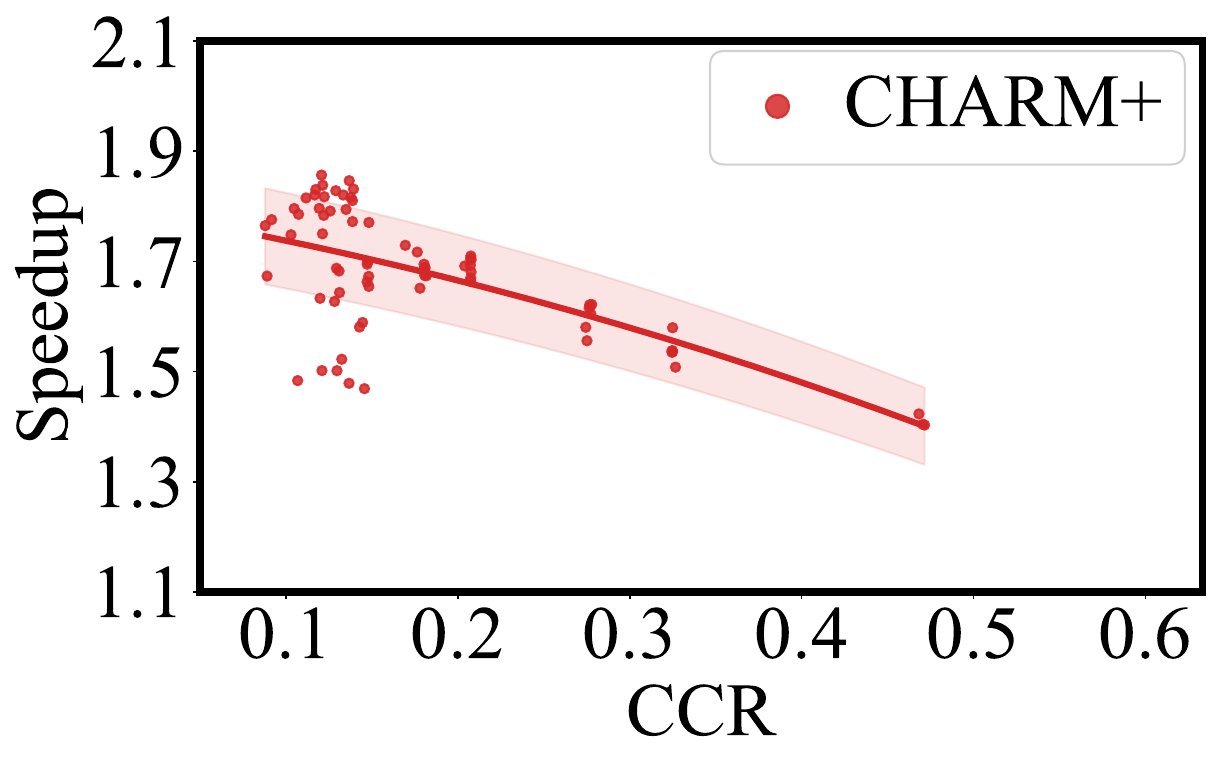}
    \end{minipage}\hfill
    \begin{minipage}{0.32\textwidth}
        \centering
        {\textbf{Forward + Backward}\par}
        \includegraphics[width=\linewidth]{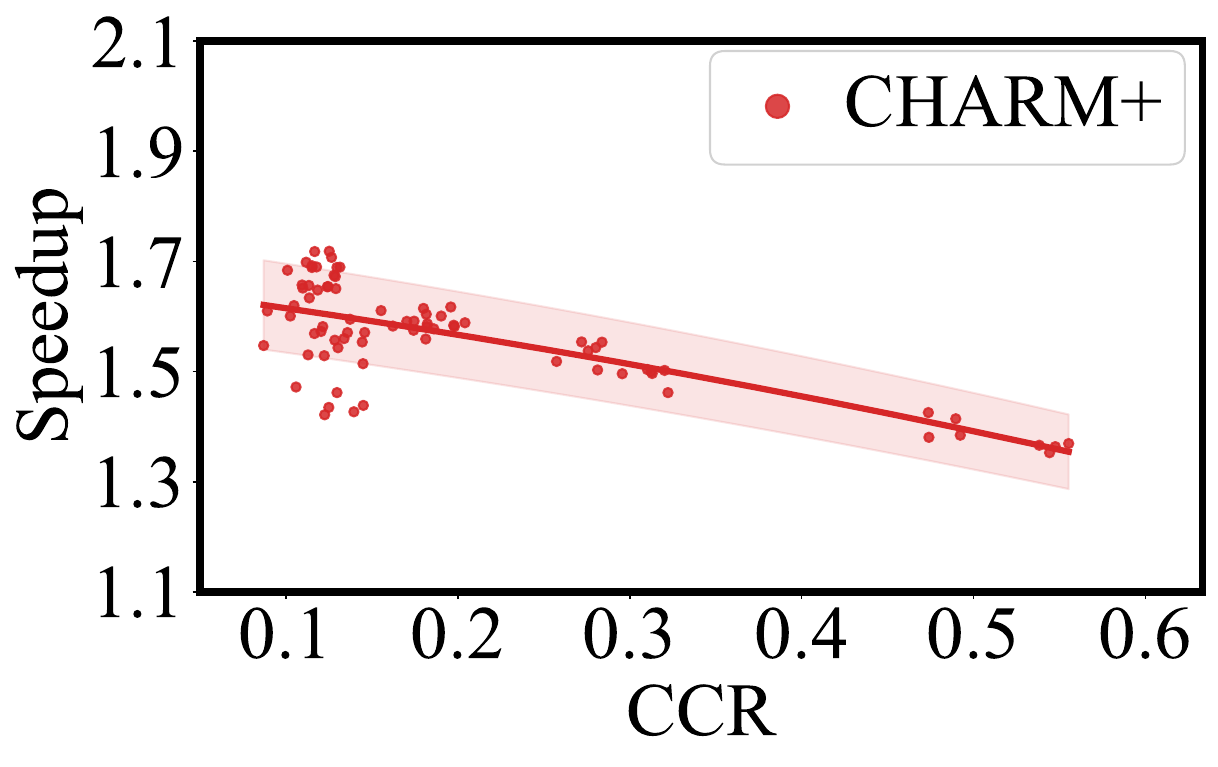}
    \end{minipage}
    \vspace{-0.4em}
    \subcaptionbox{$8\times$L20 GPU }
    {\hspace{0.9\linewidth}}

    \vspace{0.4em}

    \begin{minipage}{0.33\textwidth}
        \centering
        \includegraphics[width=\linewidth]{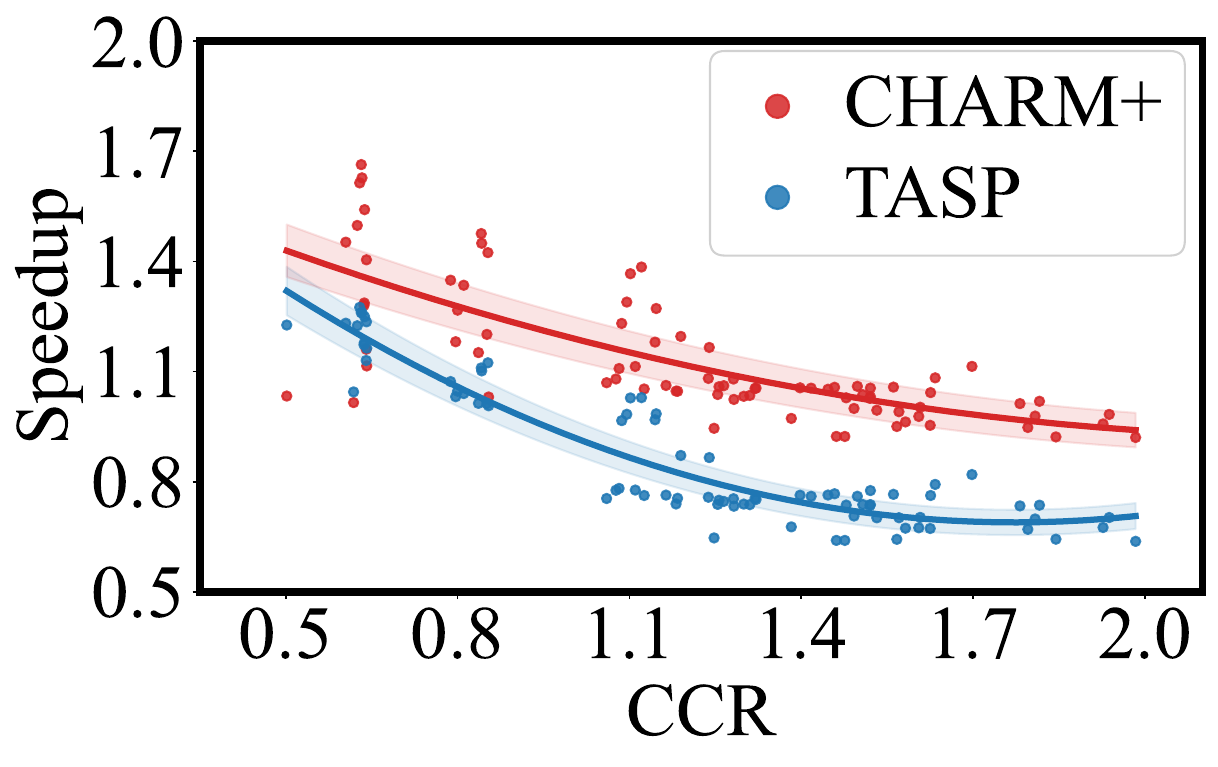}
    \end{minipage}\hfill
    \begin{minipage}{0.33\textwidth}
        \centering
        \includegraphics[width=\linewidth]{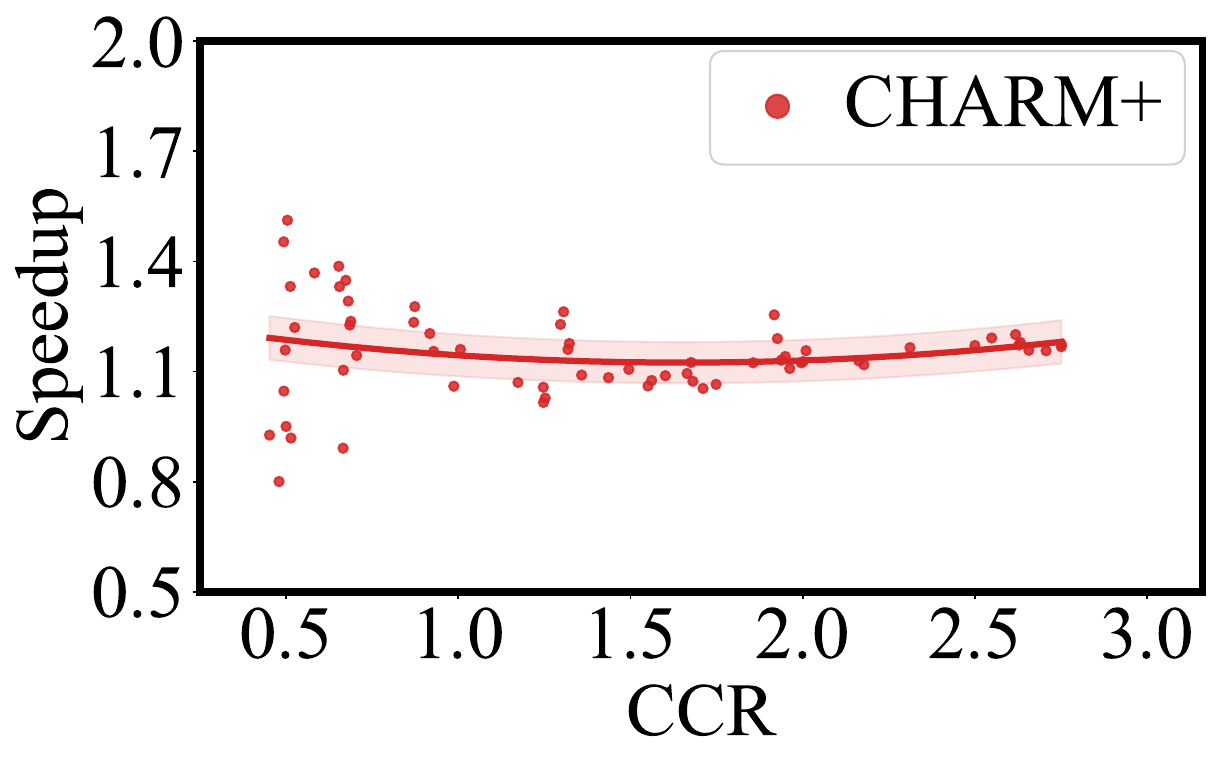}
    \end{minipage}\hfill
    \begin{minipage}{0.33\textwidth}
        \centering
        \includegraphics[width=\linewidth]{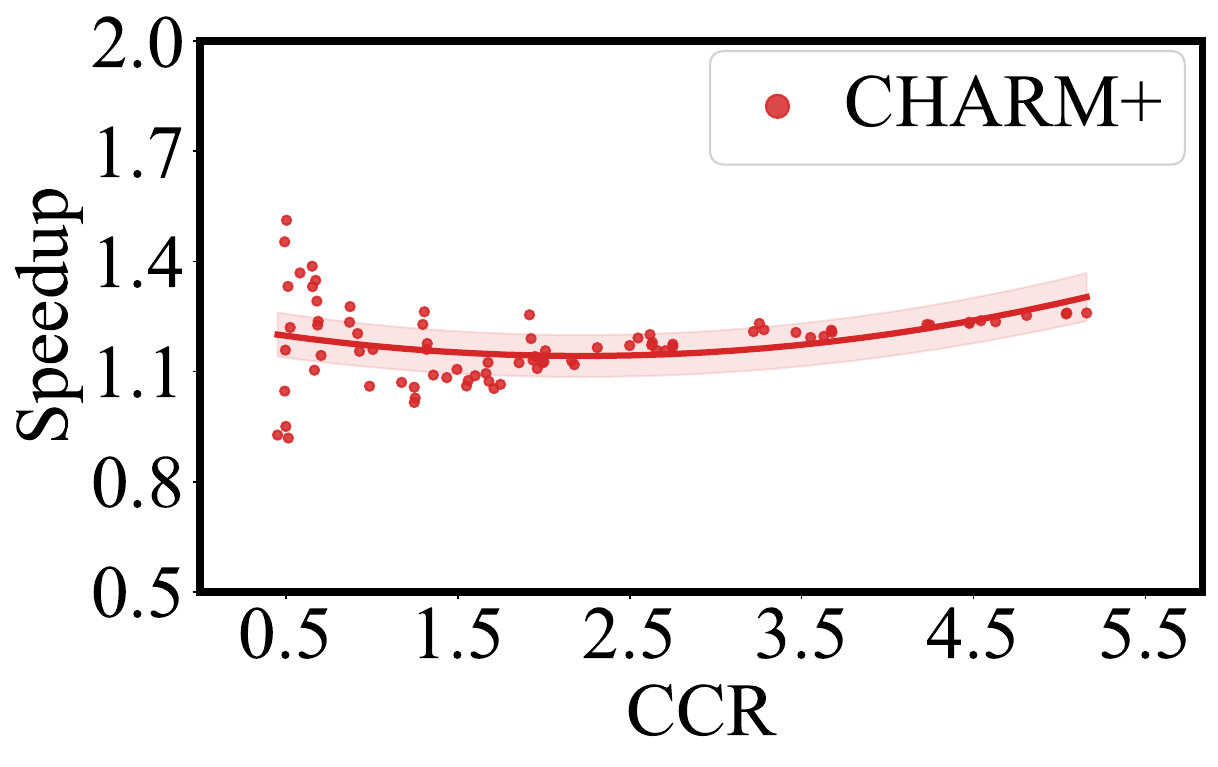}
    \end{minipage}
    \vspace{-0.4em}
    \subcaptionbox{$8\times$H20 GPU}{\hspace{0.9\linewidth}}
\caption{Single-node Experiments on $8\times$L20 (PCIe) and $8\times$H20 (NVLink). Speedup is normalized to Ring-Attention, comparing forward, backward, and end-to-end time. In different architectures, our method consistently outperforms the baseline and TASP.}
\vspace{-0.4cm}
\label{fig:SingleNodeExp}
\end{figure*}
\begin{figure*}[t]
    \centering
    \begin{minipage}{0.33\textwidth}
        \centering
        {$\mathbf{2\times4}$ \textbf{H20}\par} 
        \includegraphics[width=\linewidth]{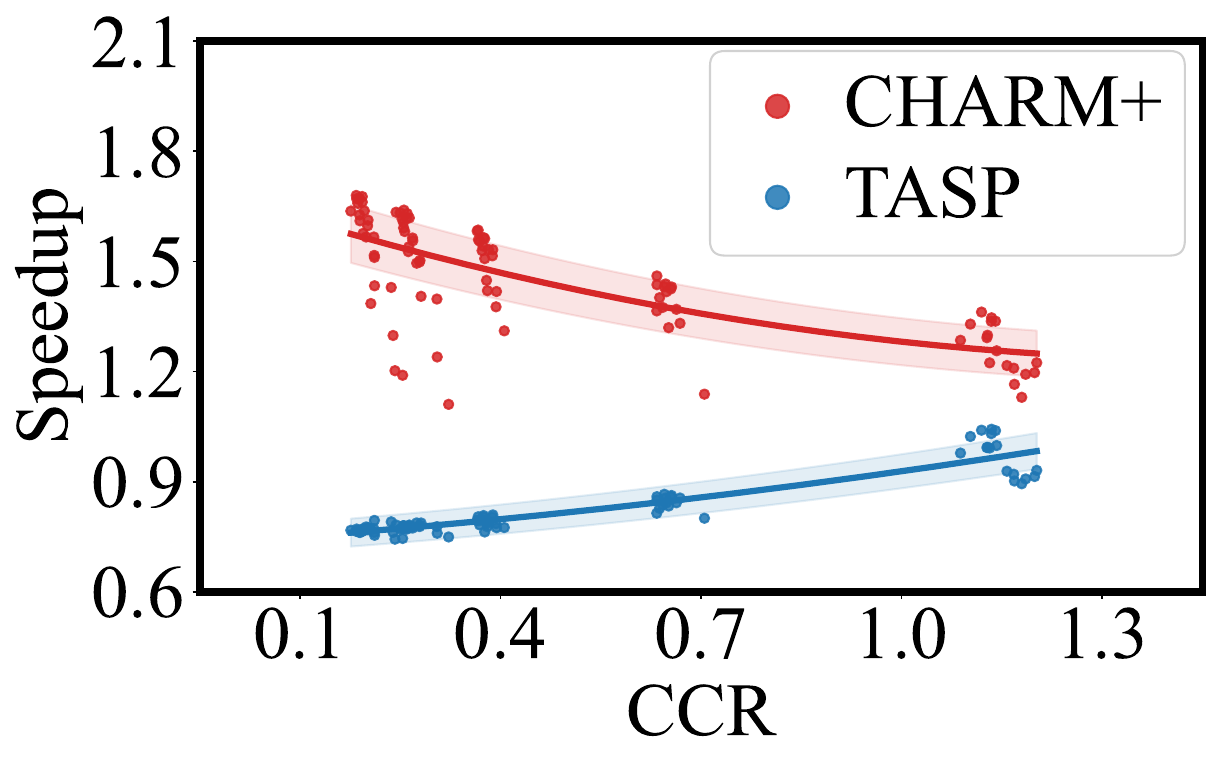}
    \end{minipage}\hfill
    \begin{minipage}{0.33\textwidth}
        \centering
        {$\mathbf{2\times8}$ \textbf{H20}\par}
        \includegraphics[width=\linewidth]{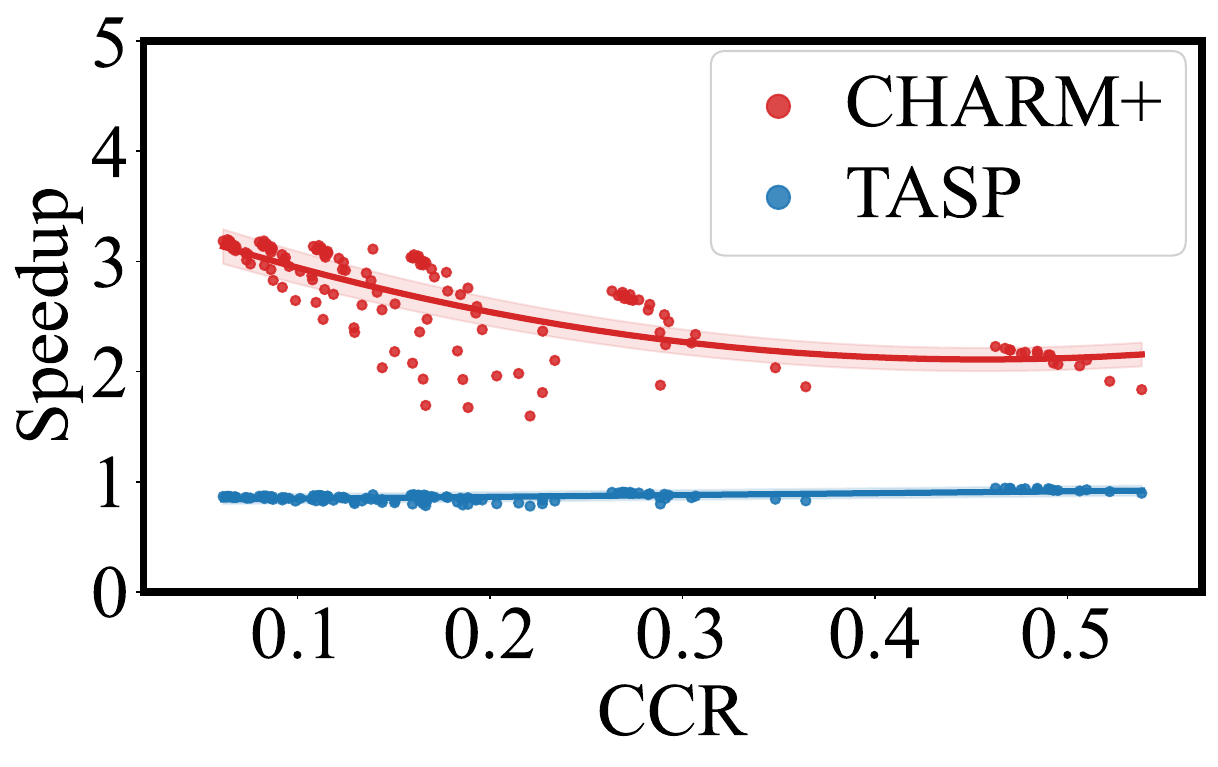}
    \end{minipage}\hfill
    \begin{minipage}{0.33\textwidth}
        \centering
        {$\mathbf{2\times8}$ \textbf{H20 Zigzag}\par}
        \includegraphics[width=\linewidth]{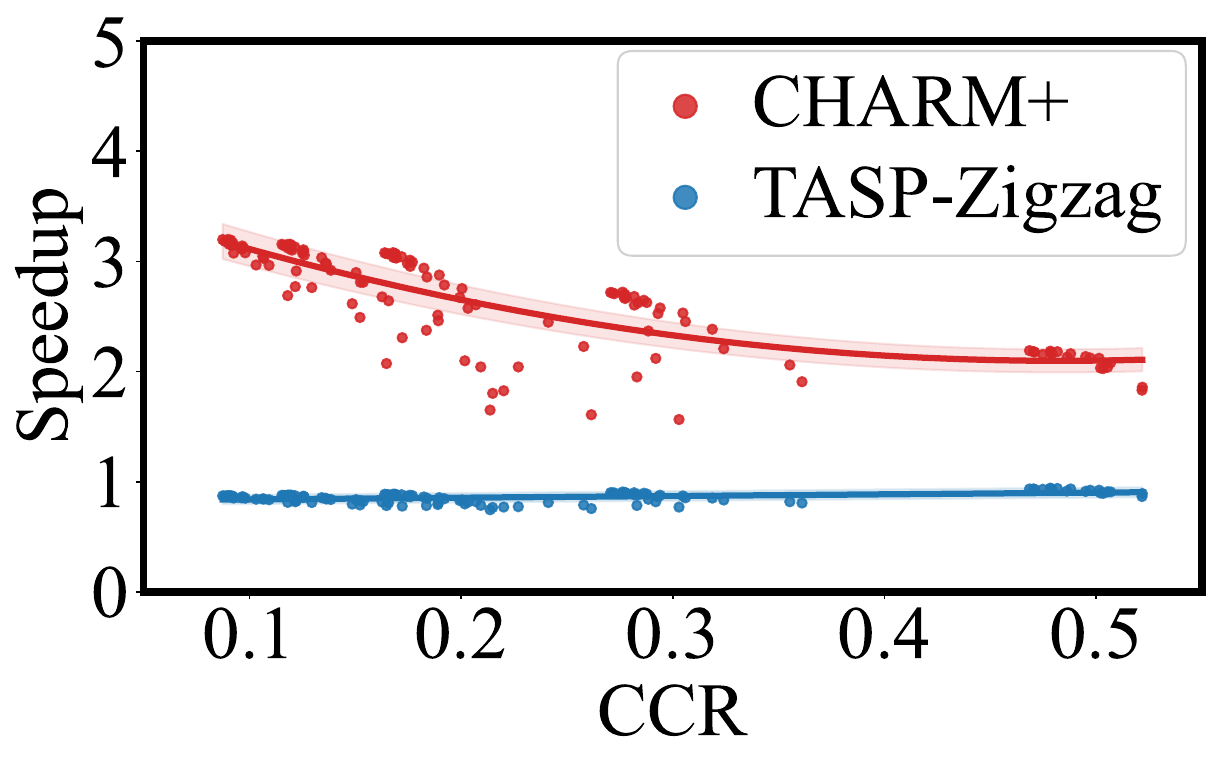}
    \end{minipage}

    \begin{minipage}{0.33\textwidth}
        \centering
        \includegraphics[width=\linewidth]{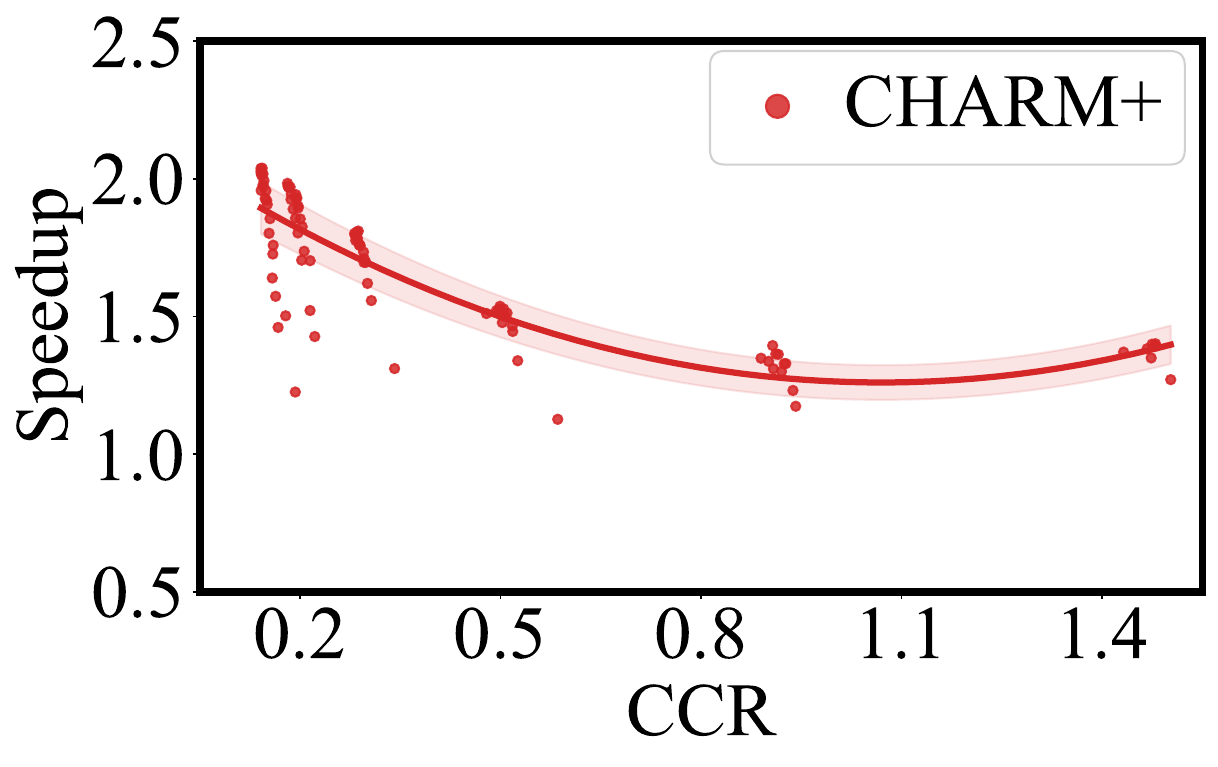}
    \end{minipage}\hfill
    \begin{minipage}{0.33\textwidth}
        \centering
        \includegraphics[width=\linewidth]{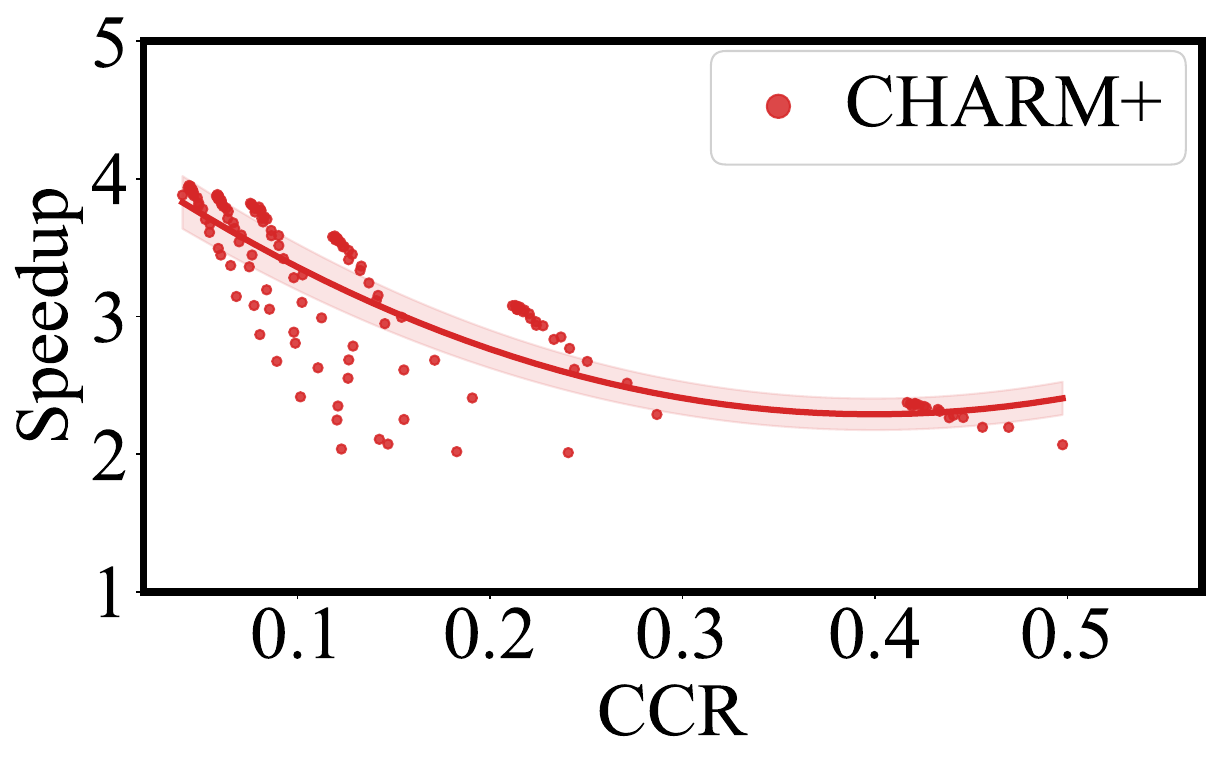}
    \end{minipage}\hfill
    \begin{minipage}{0.33\textwidth}
        \centering
        \includegraphics[width=\linewidth]{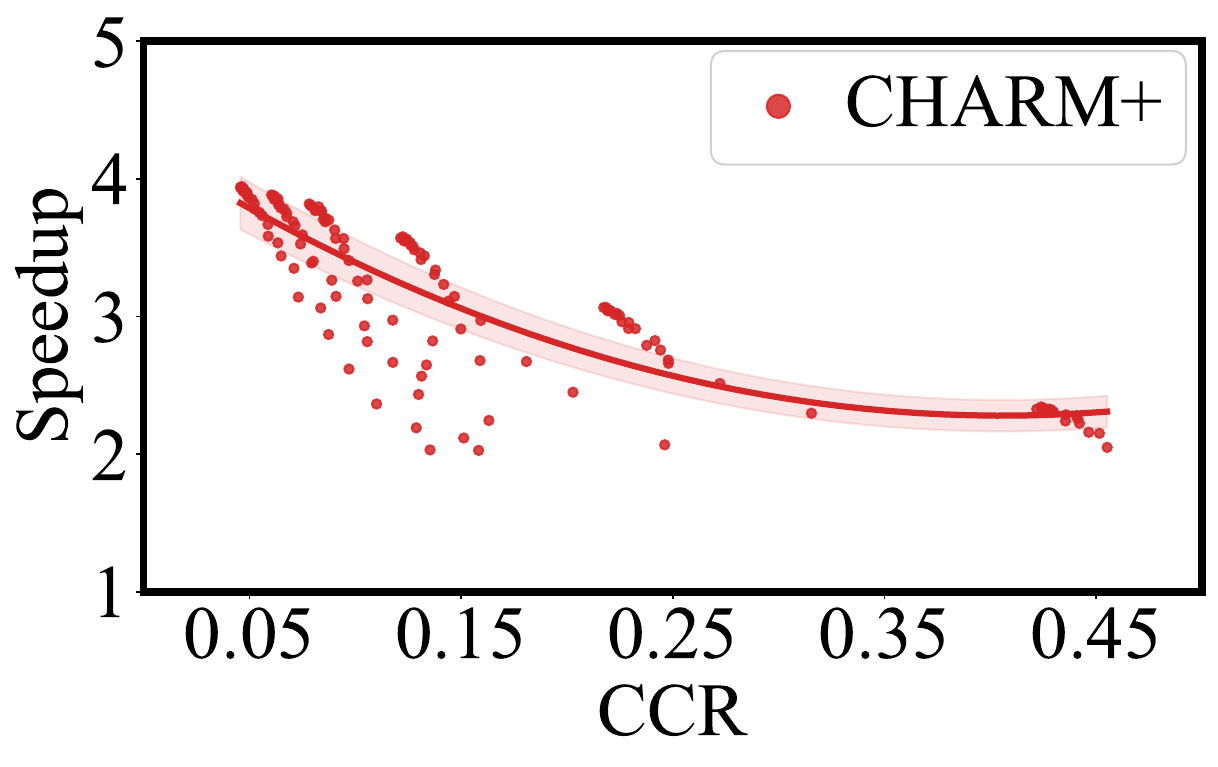}
    \end{minipage}
\caption{Multi-node Experiments on 8×H20 and 16×H20 (NVLink). Speedup is normalized to Ring-Attention and Ring-Zigzag. In the forward (top) and the backward (bottom) comparison, our method achieves consistently superior performance. Particularly, the 2×4 and 2×8 experiments highlight that our method's performance grows as the number of GPUs increases.}
\vspace{-0.6cm}
\label{fig:MultiNodeExp}
\end{figure*}
\subsection{Single-node Evaluation}

As illustrated in \cref{fig:SingleNodeExp}, we perform single-node experiments on L20 with 8 GPUs under a dual-NUMA PCIe architecture and on H20 with 8 GPUs using NVLink. Our method outperforms TASP and Ring-Attention, even in the presence of link contention (\cref{fig:SingleNodeExp}a). The poor performance of TASP on L20 is mainly attributed to its all-to-all collective communication, which fully utilizes all links and induces severe contention. Furthermore, in the H20 experiments, the observed speedup is slightly lower. This is due to the very high intra-node bandwidth (as reported in \cref{tab:main_experiment_setup}), which increases the CCR relative to other experiments. In other words, computation outweighs communication, limiting the benefits of the communication-computation overlap.

\begin{table}[t]
\centering
\caption{Memory Overhead. By setting various parameter scales, we compare the memory overhead with Ring-Attention. Our method shows comparable overhead to Ring-Attention.
}
\vspace{0.1cm}
\renewcommand{\arraystretch}{1}
\resizebox{0.95\columnwidth}{!}{
\begin{tabular}{c c c| c c||c c c| c c}
\toprule
Batch & Seq & \multirow{2}{*}{Heads} & Ring & CHARM+ 
& Batch & Seq & \multirow{2}{*}{Heads} & Ring & CHARM+ \\
Size  & Len &  & (MB) & (MB)
& Size & Len &  & (MB) & (MB) \\
\midrule

16  & 1024  & 32 & 2,179 & 2,215 
& 2 & 53,760 & 12 & 13,123 & 13,467 \\

16  & 1024  & 64 & 3,701 & 3,881 
& 4 & 53,760 & 12 & 19,301 & 20,565 \\

64  & 8192  & 32 & 45,079 & 45,423 
& 2 & 26,880 & 12 & 10,745 & 10,803 \\

128 & 8192  & 16 & 44,720 & 45,432 
& 4 & 26,880 & 12 & 14,029 & 14,425 \\

\bottomrule
\end{tabular}
}
\vspace{-0.7cm}
\label{tab:memory_assumption}
\end{table}
\textbf{Memory Overhead Analysis.} We evaluate memory overhead on L20 with 8 GPUs across varying sequence lengths for both forward and backward passes, comparing Ring-Attention with our method in \cref{tab:memory_assumption}. The results indicate that our method incurs only a modest increase in memory consumption. Furthermore, as the number of additional buffers introduced by our approach is constant (i.e., $O(1)$), memory overhead does not scale with the number of GPUs. See \cref{sec:memory_overhead} for detail analysis.

\textbf{Numerical correctness verification.} We evaluate the numerical correctness of CHARM+ by comparing its outputs and gradients against FlashAttention. The results show that CHARM+ exhibits numerical differences of the same order of magnitude as Ring-Attention across the output, log-sum-exp (LSE), and Q/K/V gradients, indicating that CHARM+ introduces no substantial additional numerical error. Detailed experimental configurations and results are provided in \cref{sec:numerical_correctness}.

\subsection{Multi-node Evaluation}

We conducted multi-node experiments on the H20 system with 8 or 16 GPUs using NVLink. On the 16-GPU setup, we compare our method 
against Ring-Attention, TASP, and their load-balanced variants, Ring-Zigzag and TASP-Zigzag. 
As shown in \cref{fig:MultiNodeExp}, our method shows a clear advantage\begin{wraptable}{r}{0.5\textwidth}
\centering
\vspace{-0.2cm}
\caption{Ablation Study. We compare the forward performance of Ring-Attention, RM+, and CHARM+ across various GPU configurations and sequence lengths. Notably,  CHARM+ will be evaluated only in cross-hardware configurations.}
\label{tab:multi_gpu_ablation}
\resizebox{0.5\columnwidth}{!}{
\begin{tabular}{c c c c c c}
\toprule
\multirow{2}{*}{GPUs} &
\multirow{2}{*}{Cross} &
\multirow{2}{*}{Seq} &
\multicolumn{3}{c}{Forward Time (ms)} \\
\cmidrule(lr){4-6}
 &  &  & Ring & RM+ & Ours \\
\midrule
\multirow{2}{*}{4 $\times$ L20} 
 & \multirow{2}{*}{\xmark}
 & 1792 & \textbf{9.35} & {10.29} & -- \\
 &  & 7168 & \textbf{36.86} & {39.66} & -- \\
\midrule
\multirow{2}{*}{8 $\times$ L20}
 & \multirow{2}{*}{\cmark}
 & 1792 & 17.61 & {25.91} & \textbf{14.81} \\
 &  & 7168 & 69.85 &{98.49} & \textbf{52.40} \\
\midrule
\multirow{2}{*}{4 $\times$ H20}
 & \multirow{2}{*}{\xmark}
 & 1792 & 5.38 & \textbf{4.76} & -- \\
 &  & 7168 & 38.62 & \textbf{33.20} & -- \\
\midrule
\multirow{2}{*}{8 $\times$ H20}
 & \multirow{2}{*}{\xmark}
 & 1792 & 5.51 & \textbf{3.65} & -- \\
 &  & 7168 & 26.79 & \textbf{13.84} & -- \\
\midrule
\multirow{2}{*}{16 $\times$ H20}
 & \multirow{2}{*}{\cmark}
 & 1792 & 32.17 & {39.04} & \textbf{12.16} \\
 &  & 7168 & 84.58 & {98.98} & \textbf{28.91} \\
\bottomrule
\end{tabular}
}
\vspace{-1.3cm}
\end{wraptable}
in cross-node configurations. 
This is because cross-node bandwidth is significantly lower than intra-node bandwidth, as listed in \cref{tab:main_experiment_setup}. 
Consequently, Ring-Attention and TASP incur GPU idle time due to frequent synchronized communication, resulting in a performance bottleneck that degrades efficiency.

\textbf{Impact of Scaling on Performance.} 
To evaluate scalability, we conducted experiments on the H20 system with 2 nodes using 8 GPUs and 2 nodes using 16 GPUs. 
As shown in \cref{fig:MultiNodeExp}, our method achieves substantial speedups at larger scales by leveraging additional GPUs for better parallelism and effectively overlapping the communication with the computation.

\subsection{Ablation Study}
\label{sec:ablation_study}

\textbf{The effectiveness of the Re-Merge Multistream Mechanism (RM+).}  We evaluate RM+ across sequence lengths on a single 8-GPU H20 node to isolate its effectiveness from link contention and cross-node communication.
As shown in \cref{tab:multi_gpu_ablation}, RM+ consistently outperforms Ring-Attention. This improvement stems from the parallel communication enabled by multiple independent $q$ and $kv$ streams. As a result, RM+ achieves a maximum speedup of $1.9\times$ over Ring-Attention.

\textbf{The effectiveness of Cross-Hardware Query Parallelism.} We compare CHARM+ against Ring-Attention and RM+ in both cross-NUMA and cross-node configurations. Specifically, for cross-NUMA experiments, we use 8 GPUs on the L20 system, while for cross-node experiments, we use 16 GPUs on the H20 system. As shown in \cref{tab:multi_gpu_ablation}, RM+ suffers significant performance degradation in cross-hardware configurations, whereas CHARM+ achieves substantial improvements, with up to 1.9× and 3.5× speedup in cross-NUMA and cross-node configurations, respectively. The performance gain is smaller on L20 than on H20 due to link contention on L20, as discussed in the next subsection. The performance drop of RM+ is primarily due to a bandwidth imbalance between intra- and cross-hardware communication, as well as increased cross-node communication resulting from multistream parallelism. In contrast, CHARM+ decouples inter-node and intra-node communication to reduce synchronization overhead and improve overall performance.
\begin{table*}[t]
\setlength{\tabcolsep}{4.5pt}
\centering
\caption{Performance on H100. Note that TASP requires the sequence length to be divisible by $n(n-1)$. Therefore, the sequence lengths for the four-node and two-node settings differ slightly.}
\label{tab:H100_experiments}
\resizebox{0.95\linewidth}{!}{%
\begin{tabular}{c c cccc cccc}
\toprule

\multirow{2}{*}{B} &
\multirow{2}{*}{Seq} &
\multicolumn{4}{c}{Forward} &
\multicolumn{4}{c}{Backward} \\

\cmidrule(lr){3-6}
\cmidrule(lr){7-10}

& &
Ring & TASP & CHARM+ & Speedup &
Ring & TASP & CHARM+ & Speedup \\

\midrule

\multicolumn{10}{c}{\textbf{GPUs = $\boldsymbol{2\times 8}$}} \\
\midrule

16  & 26,880 & 149.97 & 98.01 & \textbf{63.93} & \textbf{2.35$\times$}
                & 449.58 & -- & \textbf{141.52} & \textbf{3.18$\times$} \\

16  & 13,440 & 72.65 & 49.27 & \textbf{31.84} & \textbf{2.28$\times$}
                & 226.46 & -- & \textbf{65.35} & \textbf{3.47$\times$} \\

64  & 6,720 & 141.75 & 97.65 & \textbf{61.80} & \textbf{2.29$\times$}
               & 448.84 & -- & \textbf{123.32} & \textbf{3.64$\times$} \\

128 & 3,360 & 140.55 & 97.15 & \textbf{61.54} & \textbf{2.28$\times$}
               & 450.01 & -- & \textbf{121.99} & \textbf{3.69$\times$} \\

\midrule

\multicolumn{10}{c}{\textbf{GPUs = $\boldsymbol{4\times 8}$}} \\
\midrule

16  & 31,744 & 185.42 & 183.22 & \textbf{68.17} & \textbf{2.72$\times$}
                & 559.34 & -- & \textbf{146.01} & \textbf{3.83$\times$} \\

16  & 15,872 & 89.27 & 93.30 & \textbf{34.64} & \textbf{2.58$\times$}
                & 279.85 & -- & \textbf{70.62} & \textbf{3.96$\times$} \\

64  & 7,936 & 179.13 & 182.93 & \textbf{66.48} & \textbf{2.69$\times$}
               & 556.25 & -- & \textbf{134.48} & \textbf{4.14$\times$} \\

128 & 3,968 & 178.51 & 183.09 & \textbf{66.43} & \textbf{2.69$\times$}
               & 560.63 & -- & \textbf{133.05} & \textbf{4.21$\times$} \\

\bottomrule
\end{tabular}
}
\vspace{-0.4cm}
\end{table*}

\textbf{The impact of link contention in PCIe architectures.} 
To avoid communication synchronization issues caused by cross-NUMA communication on the L20 system, we conducted single-NUMA experiments on an L20 node with 4 GPUs, and performed comparison experiments on an H20 node with 4 GPUs connected via NVLink, which eliminates link contention. 
\subsection{Evaluation on High-end GPU}
\label{sec:H100}
As shown in \cref{tab:multi_gpu_ablation}, on L20, RM+ suffers significant performance degradation due to link contention, performing worse than Ring-Attention. In contrast, RM+ exhibits much better performance on H20.
\begin{wrapfigure}{l}{0.55\textwidth} 
\vspace{-0.3cm}
    \centering
    \includegraphics[width=0.48\columnwidth]{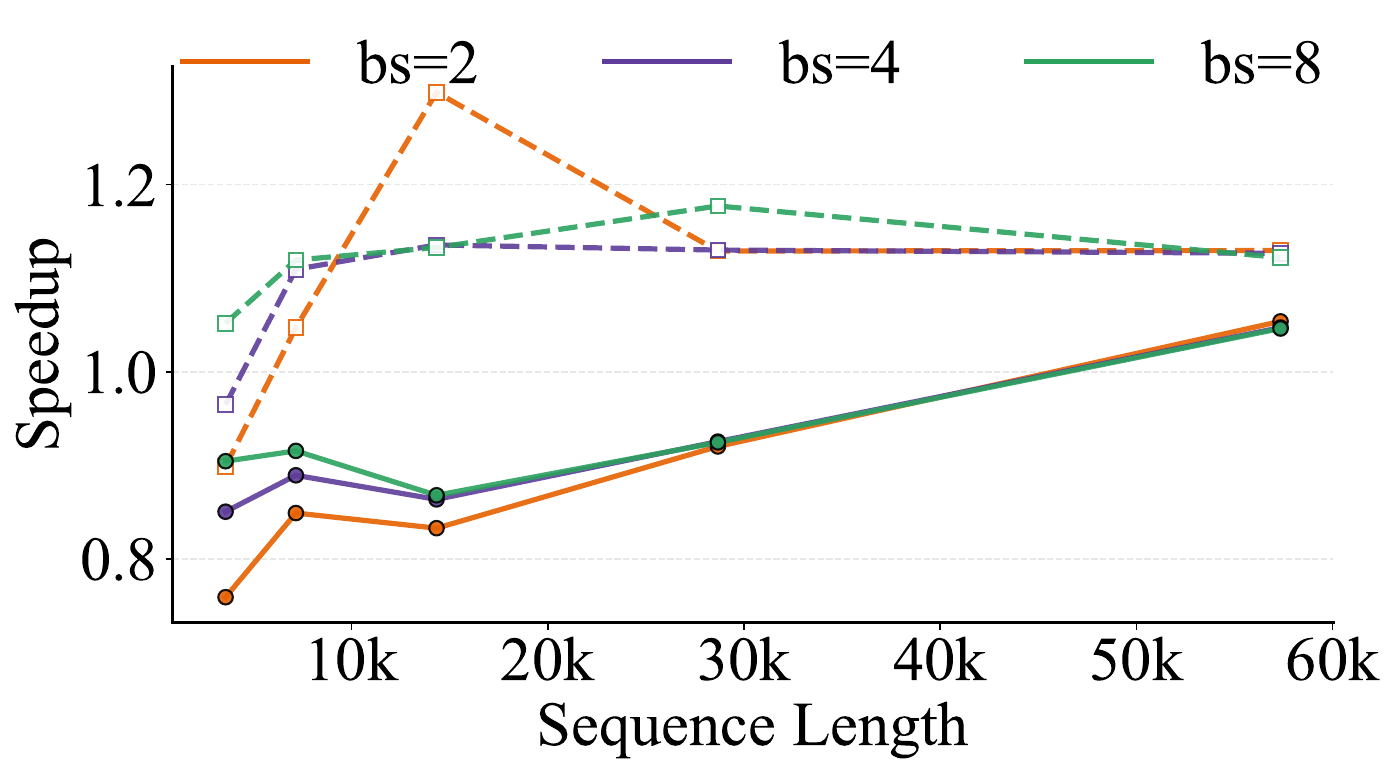}
    \caption{Performance of RM+ under PCIe and NVLink. We evaluate the forward time of RM+ on $4\times$H20 GPUs (dashed) and $4\times$L20 GPUs (solid) across different batch sizes (bs). Compared to Ring-Attention, performance on L20 is lower than on H20 due to the link contention.
}
    \vspace{-0.8cm}
\label{fig:ablation_pcie_contention}
\end{wrapfigure}
We further conduct a comprehensive evaluation with various batch sizes and sequence lengths, summarized in \cref{fig:ablation_pcie_contention}. 
The results indicate that link contention degrades performance in almost all configurations. 
The exception occurs when both the batch size and sequence length are small. 

In this case, the bandwidth is not fully utilized, so the link contention is negligible. Meanwhile, it is worth noting that in the $8\times$ L20 experiments, link contention leads to smaller performance gains than on H20. Nevertheless, enabled by Cross-Hardware Query Parallelism, CHARM+ still significantly outperforms Ring-Attention and TASP, as shown in \cref{fig:SingleNodeExp}a.

To completely validate our method on the high-end GPUs, we rented 32 H100 GPUs equipped with high-bandwidth NVLink. We conducted full forward and backward experiments under both 2×8 and 4×8 configurations, and the results are reported in \cref{tab:H100_experiments} (detailed results can be found in \cref{sec:scaling_experiment}). When TASP begins to leverage high-performance GPUs, our method still maintains a leading performance, achieving an average speedup of 2× in the forward and 3× in the backward compared to ring-attention. Additionally, as described in \cref{sec:3_3}, our method consistently makes intermediate cross-node communication effectively overlapped by the intra-node pipeline. In contrast, other approaches suffer from synchronization constraints, making them susceptible to the slower cross-node communication. Consequently, our method exhibits greater performance as the scale increases (speedups of $2.28\times$/$3.69\times$ on 16 H100 GPUs and $2.69\times$/$4.21\times$ on 32 H100 GPUs at maximum performance), highlighting the advantage of our method.

\subsection{End-to-end Training Performance}

\begin{table*}[t]
\centering
\caption{End-to-end Performance Comparison of Ring-Attention and CHARM+ with Model Scaling}
\label{tab:megetron_speedup_L20}
\renewcommand{\arraystretch}{1.12}
\setlength{\tabcolsep}{0pt}
\resizebox{0.95\linewidth}{!}{%
\begin{tabular*}{\textwidth}{
  @{\extracolsep{\fill}}
  c c c c c c c c
  @{}
}
\toprule
\multirow{2}{*}{Batch Size}
& \multirow{2}{*}{Seq Len}
& \multirow{2}{*}{Model}
& \multicolumn{2}{c}{Ring}
& \multicolumn{2}{c}{CHARM+}
& \multirow{2}{*}{Speedup} \\
\cmidrule(lr){4-5} \cmidrule(lr){6-7}
& & & Latency & Memory & Latency & Memory & \\
\midrule
16 & 2,048  & GPT-350M & 10082.6 & 11,372 & \textbf{4847.0}  & 11,518 & \textbf{2.08$\times$} \\
8  & 4,096  & GPT-350M & 10032.6 & 11,348 & \textbf{4871.5}  & 11,544 & \textbf{2.06$\times$} \\
4  & 8,192  & GPT-350M & 9914.5  & 11,436 & \textbf{5173.6}  & 11,552 & \textbf{1.92$\times$} \\
2  & 16,384 & GPT-350M & 10089.5 & 12,112 & \textbf{5637.3}  & 12,170 & \textbf{1.79$\times$} \\
\midrule
16 & 2,048  & GPT-1.3B & 21505.5 & 30,816 & \textbf{11738.3} & 31,602 & \textbf{1.83$\times$} \\
8  & 4,096  & GPT-1.3B & 22073.7 & 31,190 & \textbf{11768.1} & 31,546 & \textbf{1.88$\times$} \\
4  & 8,192  & GPT-1.3B & 21857.8 & 31,414 & \textbf{12097.5} & 32,030 & \textbf{1.81$\times$} \\
2  & 16,384 & GPT-1.3B & 22064.5 & 32,270 & \textbf{12796.6} & 32,312 & \textbf{1.72$\times$} \\
\bottomrule
\end{tabular*}
}
\vspace{-0.4cm}
\end{table*}

\textbf{High-end GPU.} We integrate our operator into Megatron-LM and compare it with Megatron's ring implementation. The experimental results are shown in \cref{tab:main_megatron_speedup}. We use context parallelism (CP) for training on 16 high-end H20/H100 GPUs and report the average per-iteration time in milliseconds.

\begin{table}[htbp]
\vspace{-0.3cm}
\centering
\caption{End-to-end Performance Comparison of Ring-Attention and CHARM+ with High-end GPU}
\label{tab:main_megatron_speedup}
\vspace{0.1cm}
\resizebox{0.95\linewidth}{!}{%
\begin{tabular}{c c c cc cc cc}
\toprule
\multirow{2}{*}{Batch Size} & \multirow{2}{*}{Seq Len} & \multirow{2}{*}{Model} 
& \multicolumn{2}{c}{Ring} 
& \multicolumn{2}{c}{CHARM+} 
& \multicolumn{2}{c}{Speedup} \\
\cmidrule(lr){4-5} \cmidrule(lr){6-7} \cmidrule(lr){8-9}
& & & H20 & H100 & H20 & H100 & H20 & H100 \\
\midrule
16  & 16,384 & GPT3-350M & 7304.6 & 3385.1 & \textbf{5601.3} & \textbf{1925.7} & \textbf{1.30$\times$} & \textbf{1.76$\times$} \\
32  & 8,192  & GPT3-350M & 7193.7 & 3291.5 & \textbf{4669.7} & \textbf{1679.2} & \textbf{1.54$\times$} & \textbf{1.96$\times$} \\
64  & 4,096  & GPT3-350M & 7169.6 & 3234.6 & \textbf{4452.5} & \textbf{1538.6} & \textbf{1.61$\times$} & \textbf{2.10$\times$} \\
128 & 2,048  & GPT3-350M & 7143.6 & 3210.8 & \textbf{4242.5} & \textbf{1482.0} & \textbf{1.68$\times$} & \textbf{2.17$\times$} \\
\bottomrule
\end{tabular}
}
\end{table}


Notably, Megatron-LM includes extensive optimizations for ring attention that our method does not incorporate. Moreover, non-attention linear layers dominate end-to-end training, limiting the impact of distributed attention optimization. Despite this, our method still significantly outperforms ring-attention on H20 and H100, achieving max $1.68\times$ and $2.17\times$ speedup, respectively. Furthermore, our method performs better at larger batch sizes, making it more suitable for real-world workloads.



\textbf{Model scaling.} To evaluate the scalability of CHARM+ across model sizes, we conduct additional end-to-end experiments on 16 L20 GPUs with model sizes of 350M and 1.3B, measuring iteration latency (ms) and GPU memory consumption (MB). Although model configurations affect the relative computational costs of attention and MLP layers, they do not fundamentally change the communication patterns targeted by CHARM+. As shown in \cref{tab:megetron_speedup_L20}, CHARM+ achieves consistent speedups of $1.72\times$–$2.08\times$ across both model sizes and all tested configurations, with nearly identical memory overhead. These results demonstrate the effectiveness of CHARM+ across different model scales.

\section{Conclusion}
In this work, we present CHARM+, a novel approach that reorganizes the communication process of Ring-Attention using a merge-based strategy. Furthermore, it leverages a multistream parallelism of $q$ and $kv$ to reduce communication dependencies and decouples cross-hardware and intra-hardware communication. We implement both the forward and backward algorithms of the mechanism and conduct extensive experiments and ablation studies across various settings. Experiment results across diverse hardware platforms and GPUs with varying computational throughput demonstrate that CHARM+ consistently achieves significant performance improvements over state-of-the-art methods.




\bibliography{references}
\bibliographystyle{abbrv}

\medskip






\appendix

\newpage
\section*{Appendix}

\section{Implementation}
We implement both the forward and backward of CHARM+. \cref{alg:build_groups} presents the logic for constructing the communication groups used in our method. Specifically, the communication peers are partitioned into two disjoint subgroups corresponding to $kv$ communication and $q$ communication at each merge level. To avoid imbalanced communication overhead where multiple GPUs simultaneously initiate communication to the same GPU, we determine the order of $kv$ and $q$ communication based on each GPU’s relative position within the global communication group. For example, when GPU 0 and 1 communicate with GPU 2 and 3, GPU 0 sends $kv$ to GPU 2 and $q$ to GPU 3, whereas GPU 1 sends $q$ to GPU 2 and $kv$ to GPU 3. This complementary scheduling ensures that both the $kv$ and $q$ processing streams on each GPU remain active, effectively preventing stream idling and improving overall communication efficiency.

\begin{algorithm}[htbp]
  \caption{GenerateCommList}
  \label{alg:build_groups}
  \begin{algorithmic}
    \STATE {\bfseries Input:} processGroup $pg$
    \STATE {\bfseries Output:} $\text{commList}_{q}$, $\text{commList}_{kv}$
    \STATE Initialize $tmp\leftarrow[], n\leftarrow\text{len}(pg), x\leftarrow\text{getRank}(pg)$
    \STATE $l \leftarrow 1$
    \WHILE{$l < n$}
      \STATE $g \leftarrow \{x \oplus i \mid i \in [l, 2 \cdot l)\}$
      \STATE Append $g$ to $tmp$
      \STATE $l \leftarrow 2 \cdot l$
    \ENDWHILE

    \STATE Initialize $g_{1},g_{2}\leftarrow[]$, append $tmp[0]$ to $g_{1}$
    \STATE $g_{3} \leftarrow \text{Sort}(tmp[0] \cup \{x\})$

    \FOR{each $g$ in $tmp[1:]$}
      \STATE $\text{first},\text{second}\leftarrow\text{split}(g)$, $\text{index}\leftarrow \text{getPosition}(x,g_{3})$
      \STATE $\text{half} \leftarrow |g_{3}| / 2$
      \IF{$\text{index} < \text{half}$}
        \STATE Append $\text{first}$ to $g_{1}$, append  $\text{second}$ to $g_{2}$
      \ELSE
        \STATE $\text{index} \leftarrow \text{index} - \text{half}$
        \STATE Append  $\text{second}$  to $g_{1}$, append  $\text{first}$  to $g_{2}$
      \ENDIF
      \STATE $g_[3] \leftarrow \text{Sort}(g_{3} \cup g)$
    \ENDFOR

    \STATE $\text{commList}_{kv},\text{commList}_{q}\leftarrow\text{concatenate}(g_{1},g_{2})$
    \STATE {\bfseries Return} $\text{commList}_{kv}$, $\text{commList}_{q}$
  \end{algorithmic}
\end{algorithm}
\begin{algorithm}[tb]
  \caption{CHARM+ Forward}
  \label{alg:forward}
  \begin{algorithmic}
    \STATE {\bfseries Input:} tensor $q,k,v$, processGroup $pg$
    \STATE {\bfseries Output:} $out$
    \STATE $\text{commList}_{q},\text{commList}_{kv} \leftarrow \text{GenerateCommList}(pg)$
    \STATE
    $q_\text{cross}\leftarrow \text{CrossComm}(\text{q})$
    \STATE Initialize $out,out_{\text{cross}} \leftarrow None, n\leftarrow\text{getsize}(pg)$
    \STATE Commwait()
    \FOR{$i\leftarrow 1$ {\bfseries to} $\log(n)$}
    \STATE $k_{\text{other}},v_{\text{other}} \leftarrow \text{KVComm}(k,v)$
    \STATE
    $q_{\text{other}},q_{\text{otherC}} \leftarrow \text{QComm}(q,q_{\text{cross}})$
    \IF{$i ==1$}
    \STATE $out \leftarrow \text{CompAttention}(q,k,v,out)$
    \STATE $out_{\text{cross}} \leftarrow \text{CompAttention}(q_{\text{cross}},k,v,out)$
    \ELSE
    \STATE $out_{\text{cross}} \leftarrow \text{CompAttentionComm}(q_{\text{otherC}},k,v,out)$
    \STATE $out \leftarrow \text{CompAttentionComm}(q_{\text{other}},k,v,out)$
    \STATE $out_{\text{cross}} \leftarrow \text{CompAttention}(q_{\text{cross}},k_{\text{other}},v_{\text{other}},out)$
    \STATE $out \leftarrow \text{CompAttention}(q,k_{\text{other}},v_{\text{other}},out)$
    \ENDIF
    \STATE Commwait()
    \ENDFOR
    \STATE $out\leftarrow\text{CommAndUpdate}(out,out_{\text{cross}})$
  \end{algorithmic}
\end{algorithm}

\cref{alg:forward} provides the core logic chain of the forward algorithm  \footnote{Since the forward and backward share the same overall logic, we omit the detailed description of the backward pass for brevity.}, which is consistent with the overall pipeline in \cref{fig:overall_framework}(f). Notably, for the $q$ communication, the attention output needs to be communicated back to the original GPU after the attention computation finishes (i.e., the \emph{Out} communication in \cref{fig:overall_framework}). Therefore, we prioritize the $q$ communication stream during execution, enabling earlier output communications and effective overlap with the subsequent computation of the $kv$ communication stream. 

\section{Analysis}
We conduct a detailed analysis of the factors affecting CCR, as well as the memory usage of Ring-Attention and CHARM+.

\subsection{Analysis of CCR}
\label{sec:CCR}

As mentioned in \cref{eq:ccr}, CCR is defined by us as:
\begin{equation}
CCR=\sum_{i=1}^{N}T^{i}_{\text{comp}(q,k,v)}/\sum_{i=1}^{N}T^{i}_{\text{comm}(k,v)}
\end{equation}
The communication time depends only on the amount of data being transferred and the available bandwidth. Different on-device and off-device connections have different bandwidths. Therefore, the relation between communication time and factors can be defined as:
\begin{equation}
T_{comm}\propto \frac{BSH}{Bandwidth}
\end{equation}

And the computation time is essentially the time complexity of FlashAttention, which is
\begin{equation}
T_{comp}\propto \frac{BS^{2}H+BSH^{2}}{FLOPS}
\end{equation}
Note that, while the theoretical time remains the relation above, the actual runtime is dominated by memory IO in standard attention and both the actual FLOPS and bandwidth exhibit some variability. It can be seen that once the input size and computation pattern are fixed, the time per step is constant. Thus, the total time equals the sum of non-overlappable times along the critical path, which means reducing the number of steps (minimizing the overall critical path) directly reduces the total time. In addition, we provide some quantitative results under different hardware and input specifications, as shown in \cref{sec:scaling_experiment}. Due to fluctuations in actual time, we collected the time data in real-time through the interface provided by CUDA events.

\subsection{Analysis of Memory Overhead}
\label{sec:memory_overhead}

Since we do not change the computation pattern but instead improve the overlap between computation and communication, we focus our analysis on the major memory-consuming buffers and ignore the memory overhead of the attention computation.

We assume that the input query, key, and value are all $[b, s, h]$ (i.e., batch size, sequence length and hidden dims). For Ring-Attention, the memory overhead of the buffers is:
\begin{equation}
M_{Ring}\approx 3bsh+2bsh+1.5bsh+5bsh+3bsh+1.5bsh+2bsh+2bsh \approx 20bsh
\end{equation}
Specifically, under FP32 precision, each buffer occupies $BSH$ memory. In Ring-Attention, one buffer stores the local query block, two buffers hold the $key$ and $value$ blocks being processed, and two more buffers receive the next $KV$ blocks to overlap computation and communication. One buffer stores the accumulated $output$, and half a buffer holds intermediate FlashAttention results (usually in BF16/FP16). During backward, five buffers store $q, k, v$, the $output$, and its gradient; three buffers store total gradients of $q, k, v$; 1.5 buffers store FlashAttention gradients; and two buffers handle the next $KV$ block computation. Finally, because $KVs$ propagate across GPUs in a ring, the accumulated gradients must be sent back to the original GPU, requiring two additional buffers.

We further analyze the memory overhead of CHARM+, which can be formulated as:
\begin{equation}
\begin{aligned}
M_{\text{CHARMS+}} &\approx 3bsh + 2bsh + 1.5bsh + 5bsh + 3bsh + 1.5bsh + 2bsh + \boldsymbol{6bsh} - 2bsh \\
                   &\approx 24bsh
\end{aligned}
\end{equation}
During forward and backward, each GPU must retain its own KV and receive Q of cross-hardware, requiring three extra buffers. Moreover, FlashAttention results can be updated immediately, so the buffers used in ring-attention for gradient sending can be omitted.

Overall, our improved method logically uses four more buffers than ring-attention. However, GPU memory is temporally reused across forward and backward and buffer lifetimes do not fully overlap, the peak buffer usage remains roughly the same, resulting in only a 3\%\textasciitilde5\% increase in observed memory.

\section{Experimental Details}

\subsection{Nsight Result}
\label{sec:nsight_compare}

\begin{figure}[!htbp]
    \centering
    \begin{subfigure}[b]{\columnwidth}
        \centering
        \includegraphics[width=\columnwidth]{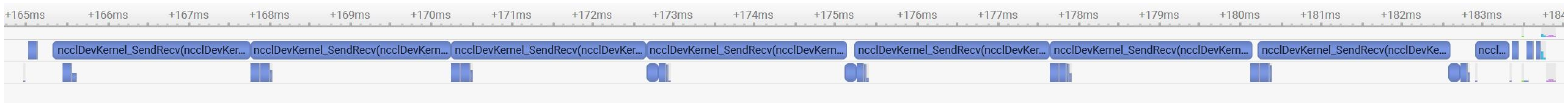}
        \caption{Ring-Attention}
        \label{fig:nsight_ring}
    \end{subfigure}

    \vspace{0.2cm}

    \begin{subfigure}[b]{\columnwidth}
        \centering
        \includegraphics[width=\columnwidth]{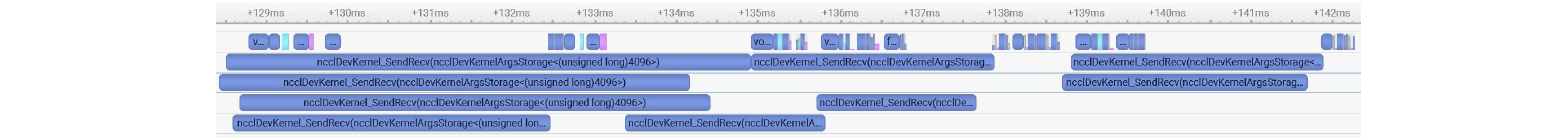}
        \caption{CHARM+}
        \label{fig:nsight_charm+}
    \end{subfigure}

    \caption{Nsight Analysis of Ring-Attention and CHARM+}
    \label{fig:nsight}
\end{figure}

To better understand the performance improvement, we profile the forward pass of Ring-Attention and CHARM+ on 8 L20 GPUs using Nsight Systems, as shown in \cref{fig:nsight_ring} and \cref{fig:nsight_charm+}, respectively. The L20 platform has relatively limited communication bandwidth, making communication kernels substantially longer than computation kernels. This imbalance makes communication–computation overlap particularly important, as the overall execution time is largely determined by how effectively communication latency can be hidden behind computation.

In \cref{fig:nsight_ring}, Ring-Attention exhibits substantial GPU idle time due to its sequential communication dependencies. Each communication step depends on the completion of the preceding step, limiting scheduling flexibility and leaving computation unable to fully hide communication latency. In contrast, as shown in \cref{fig:nsight_charm+}, CHARM+ reorganizes ring-based communication into a merge-based structure, shortening the dependency chain, and leverages multiple streams to execute independent communication and computation tasks concurrently. This design substantially reduces idle gaps and improves communication–computation overlap. Notably, both traces are plotted on the same time scale, clearly showing that CHARM+ completes execution in less time than Ring-Attention, demonstrating the effectiveness of CHARM+ in improving overall execution performance.

\subsection{Evaluation on High-end GPU}
\label{sec:scaling_experiment}
We validate our method on the high-performance H100 GPUs, with detailed configurations provided in \cref{tab:main_experiment_setup}. We conduct both forward and backward experiments under 2×8 and 4×8 setups (with a hidden dimensions of $12 \times 192$). We additionally conducted experiments on ultra-long sequences, specifically those exceeding 100k tokens. \cref{tab:forward_only_16} reports the forward results for the 2×8 configuration, \cref{tab:backward_only_16} reports the corresponding backward results, \cref{tab:forward_only_32} presents the forward results for the 4×8 configuration, and \cref{tab:backward_only_32} presents the corresponding backward results. We report the CCR as well as the detailed computation time ($T_{\text{comp}}$) and communication time ($T_{\text{comm}}$).

Our method shows better performance on 32 GPUs than on 16 GPUs. This is because the cross-node communication can be effectively overlapped with the intra-node pipeline. After decoupling via cross-hardware query parallelism, the only dependency is that cross-node query communication must finish before the intra-node pipeline begins. Accordingly, the query required for the next step can be communicated concurrently during the intra-node pipeline.

Based on this analysis and results, our method consistently confines cross-node communication to the start and end, with intermediate cross-node communication effectively overlapped by the intra-node pipeline. In contrast, other approaches suffer from synchronization constraints, making them susceptible to the slower cross-node communication. Consequently, our method exhibits greater performance as the scale increases (speedups of $2.28\times$/$3.69\times$ on 16 H100 GPUs and $2.69\times$/$4.21\times$ on 32 H100 GPUs at maximum performance), demonstrating our advantage.

In the scenario of ultra-long sequence lengths (over 200k tokens) on 16 H100 GPUs in \cref{tab:forward_only_16} and \cref{tab:backward_only_16}, the CCR exceeds 1, indicating that the computation time is greater than the communication time. At this point, computation becomes the main bottleneck, diminishing the benefits of overlap of communication and computation. Nevertheless, our method still significantly outperforms all other methods.

\subsection{Numerical correctness verification}
\label{sec:numerical_correctness}
In Ring-Attention, updates on each GPU occur in different orders, and the only precision loss comes from $\mathrm{BF16} \rightarrow \mathrm{FP32} \rightarrow \mathrm{BF16}$ conversion. This type of error exists in all distributed attention methods, and our method does not introduce any additional precision errors.

We verified numerical correctness  of our method. The results were obtained on an 8-GPU L20 setup. Specifically, we set the sequence length to 8K and generated 100 groups of randomly initialized $Q, K, V$, and output gradients, all sampled from a normal distribution. We executed FlashAttention, Ring-Attention, and CHARM+, respectively.

By performing element-wise subtraction, we computed the difference matrices between Ring-Attention and FlashAttention, as well as between CHARM+ and FlashAttention. We further evaluated these differences by calculating the mean and maximum, and finally reported the averages over the 100 runs, as summarized in \cref{tab:num_diff}.

\begin{table}[htbp]
\centering
\caption{Comparison of numerical differences between CHARM+ and Ring-Attention.}
\label{tab:num_diff}
\vspace{0.1cm}
\begin{tabular}{lcccc}
\toprule
Diff & CHARM+ (Max) & CHARM+ (Mean) & Ring (Max) & Ring (Mean) \\
\midrule
out  & 8.34147e-04 & 4.51229e-05 & 8.07648e-04 & 4.24147e-05 \\
lse  & 1.65642e-06 & 2.94155e-07 & 1.75893e-06 & 3.04284e-07 \\
dq   & 1.26290e-03 & 3.29077e-05 & 1.25977e-03 & 3.27659e-05 \\
dk   & 2.87713e-03 & 3.53419e-05 & 2.89413e-03 & 3.53193e-05 \\
dv   & 4.47177e-03 & 3.42940e-05 & 4.48120e-03 & 3.42894e-05 \\
\bottomrule
\end{tabular}
\end{table}

These results demonstrate that the numerical differences between CHARM+ and FlashAttention remain at the same order of magnitude as those between Ring-Attention and FlashAttention, indicating that CHARM+ does not introduce additional numerical errors.

\begin{table*}[t]
\centering
\caption{Forward performance comparison on 16-H100.}
\label{tab:forward_only_16}
\begin{tabular}{c c c c c c c c c}
\toprule
B & Seq & $T_{comm}$ & $T_{comp}$ & CCR 
& Ring & TASP & CHARM+ & Speedup \\
\midrule

\multicolumn{9}{c}{\textbf{Seq = 215,040}} \\
\midrule
1 & 215,040 & 81.49 & 86.01 & 1.06 & 116.76 & 86.93 & \textbf{66.74} & \textbf{1.75$\times$} \\
2 & 215,040 & 164.31 & 171.51 & 1.04 & 241.03 & 171.91 & \textbf{125.43} & \textbf{1.92$\times$} \\
\midrule

\multicolumn{9}{c}{\textbf{Seq = 107,520}} \\
\midrule
1 & 107,520 & 38.34 & 23.16 & 0.60 & 44.08 & 26.36 & \textbf{21.43} & \textbf{2.06$\times$} \\
2 & 107,520 & 74.25 & 45.20 & 0.61 & 89.20 & 51.46 & \textbf{42.28} & \textbf{2.11$\times$} \\
4 & 107,520 & 146.17 & 88.94 & 0.61 & 182.15 & 101.74 & \textbf{81.88} & \textbf{2.22$\times$} \\
\midrule

\multicolumn{9}{c}{\textbf{Seq = 53,760}} \\
\midrule
2 & 53,760 & 38.27 & 14.39 & 0.36 & 39.18 & 25.81 & \textbf{18.25} & \textbf{2.15$\times$} \\
4 & 53,760 & 74.39 & 26.50 & 0.36 & 78.95 & 50.29 & \textbf{34.36} & \textbf{2.30$\times$} \\
8 & 53,760 & 146.24 & 51.13 & 0.35 & 156.57 & 99.39 & \textbf{66.91} & \textbf{2.34$\times$} \\

\midrule
\multicolumn{9}{c}{\textbf{Seq = 26,880}} \\
\midrule
2 & 26,880 & 20.13 & 5.63 & 0.28 & 18.35 & 13.56 & \textbf{9.43} & \textbf{1.95$\times$} \\
4 & 26,880 & 38.24 & 9.87 & 0.26 & 37.90 & 25.64 & \textbf{17.03} & \textbf{2.22$\times$} \\
8 & 26,880 & 74.44 & 17.22 & 0.23 & 75.12 & 49.75 & \textbf{32.60} & \textbf{2.30$\times$} \\
16 & 26,880 & 146.53 & 32.05 & 0.22 & 149.97 & 98.01 & \textbf{63.93} & \textbf{2.35$\times$} \\

\midrule
\multicolumn{9}{c}{\textbf{Seq = 13,440}} \\
\midrule
2 & 13,440 & 10.89 & 2.92 & 0.27 & 9.15 & 8.18 & \textbf{5.89} & \textbf{1.55$\times$} \\
4 & 13,440 & 20.31 & 4.10 & 0.20 & 17.89 & 13.56 & \textbf{9.07} & \textbf{1.97$\times$} \\
8 & 13,440 & 38.31 & 6.87 & 0.18 & 35.72 & 25.48 & \textbf{16.57} & \textbf{2.16$\times$} \\
16 & 13,440 & 74.55 & 11.83 & 0.16 & 72.65 & 49.27 & \textbf{31.84} & \textbf{2.28$\times$} \\
32 & 13,440 & 145.95 & 21.79 & 0.15 & 143.05 & 97.30 & \textbf{62.34} & \textbf{2.29$\times$} \\

\midrule
\multicolumn{9}{c}{\textbf{Seq = 6,720}} \\
\midrule
4 & 6,720 & 10.92 & 2.53 & 0.23 & 9.12 & 8.19 & \textbf{5.63} & \textbf{1.62$\times$} \\
8 & 6,720 & 20.12 & 3.27 & 0.16 & 17.59 & 13.47 & \textbf{8.91} & \textbf{1.98$\times$} \\
16 & 6,720 & 38.26 & 4.61 & 0.14 & 35.76 & 25.27 & \textbf{16.31} & \textbf{2.19$\times$} \\
32 & 6,720 & 74.53 & 7.90 & 0.12 & 70.84 & 49.23 & \textbf{31.53} & \textbf{2.25$\times$} \\
64 & 6,720 & 146.14 & 17.05 & 0.11 & 141.75 & 97.65 & \textbf{61.80} & \textbf{2.29$\times$} \\

\midrule
\multicolumn{9}{c}{\textbf{Seq = 3,360}} \\
\midrule
8 & 3,360 & 10.98 & 2.55 & 0.25 & 9.17 & 8.10 & \textbf{5.62} & \textbf{1.63$\times$} \\
16 & 3,360 & 20.17 & 3.05 & 0.15 & 17.49 & 13.37 & \textbf{8.82} & \textbf{1.98$\times$} \\
32 & 3,360 & 38.23 & 4.61 & 0.12 & 34.50 & 25.13 & \textbf{16.23} & \textbf{2.13$\times$} \\
64 & 3,360 & 74.53 & 7.90 & 0.11 & 70.54 & 49.06 & \textbf{31.21} & \textbf{2.26$\times$} \\
128 & 3,360 & 146.20 & 14.36 & 0.10 & 140.55 & 97.15 & \textbf{61.54} & \textbf{2.28$\times$} \\

\bottomrule
\end{tabular}
\end{table*}

\begin{table*}[t]
\centering
\caption{Backward performance comparison on 16-H100.}
\label{tab:backward_only_16}
\begin{tabular}{c c c c c c c c c}
\toprule
B & Seq & $T_{comm}$ & $T_{comp}$ & CCR 
& Ring & TASP & CHARM+ & Speedup \\
\midrule

\multicolumn{9}{c}{\textbf{Seq = 215,040}} \\
\midrule
1 & 215,040 & 218.48 & 257.76 & 1.18 & 312.45 & -- & \textbf{207.07} & \textbf{1.51$\times$} \\
2 & 215,040 & 437.82 & 503.06 & 1.15 & 628.21 & -- & \textbf{405.90} & \textbf{1.55$\times$} \\
\midrule

\multicolumn{9}{c}{\textbf{Seq = 107,520}} \\
\midrule
1 & 107,520 & 109.45 & 68.03 & 0.62 & 114.64 & -- & \textbf{64.09} & \textbf{1.79$\times$} \\
2 & 107,520 & 218.37 & 135.44 & 0.62 & 228.70 & -- & \textbf{125.01} & \textbf{1.83$\times$} \\
4 & 107,520 & 436.10 & 264.15 & 0.61 & 451.53 & -- & \textbf{242.15} & \textbf{1.86$\times$} \\
\midrule

\multicolumn{9}{c}{\textbf{Seq = 53,760}} \\
\midrule
2 & 53,760 & 110.04 & 37.64 & 0.34 & 113.30 & -- & \textbf{45.91} & \textbf{2.47$\times$} \\
4 & 53,760 & 219.95 & 74.53 & 0.34 & 226.22 & -- & \textbf{88.55} & \textbf{2.55$\times$} \\
8 & 53,760 & 439.16 & 145.33 & 0.33 & 450.33 & -- & \textbf{173.18} & \textbf{2.60$\times$} \\

\midrule
\multicolumn{9}{c}{\textbf{Seq = 26,880}} \\
\midrule
2 & 26,880 & 56.73 & 11.47 & 0.20 & 57.03 & -- & \textbf{20.42} & \textbf{2.79$\times$} \\
4 & 26,880 & 112.12 & 22.35 & 0.20 & 112.24 & -- & \textbf{37.51} & \textbf{2.99$\times$} \\
8 & 26,880 & 221.73 & 43.86 & 0.20 & 226.13 & -- & \textbf{72.59} & \textbf{3.12$\times$} \\
16 & 26,880 & 442.86 & 86.72 & 0.20 & 449.58 & -- & \textbf{141.52} & \textbf{3.18$\times$} \\

\midrule
\multicolumn{9}{c}{\textbf{Seq = 13,440}} \\
\midrule
2 & 13,440 & 28.73 & 4.50 & 0.16 & 28.93 & -- & \textbf{10.81} & \textbf{2.67$\times$} \\
4 & 13,440 & 56.86 & 8.42 & 0.15 & 57.08 & -- & \textbf{18.48} & \textbf{3.09$\times$} \\
8 & 13,440 & 112.90 & 15.44 & 0.14 & 113.19 & -- & \textbf{33.84} & \textbf{3.34$\times$} \\
16 & 13,440 & 222.95 & 29.43 & 0.13 & 226.46 & -- & \textbf{65.35} & \textbf{3.47$\times$} \\
32 & 13,440 & 444.64 & 57.34 & 0.13 & 448.54 & -- & \textbf{128.53} & \textbf{3.49$\times$} \\

\midrule
\multicolumn{9}{c}{\textbf{Seq = 6,720}} \\
\midrule
4 & 6,720 & 28.88 & 3.23 & 0.11 & 29.11 & -- & \textbf{10.22} & \textbf{2.85$\times$} \\
8 & 6,720 & 56.98 & 5.94 & 0.10 & 56.67 & -- & \textbf{17.34} & \textbf{3.27$\times$} \\
16 & 6,720 & 112.98 & 10.90 & 0.10 & 112.72 & -- & \textbf{32.38} & \textbf{3.48$\times$} \\
32 & 6,720 & 224.56 & 20.72 & 0.09 & 224.23 & -- & \textbf{62.46} & \textbf{3.59$\times$} \\
64 & 6,720 & 445.33 & 40.32 & 0.09 & 448.84 & -- & \textbf{123.32} & \textbf{3.64$\times$} \\

\midrule
\multicolumn{9}{c}{\textbf{Seq = 3,360}} \\
\midrule
8 & 3,360 & 28.72 & 2.74 & 0.10 & 28.11 & -- & \textbf{10.04} & \textbf{2.80$\times$} \\
16 & 3,360 & 56.94 & 4.98 & 0.09 & 56.74 & -- & \textbf{17.06} & \textbf{3.33$\times$} \\
32 & 3,360 & 112.99 & 9.41 & 0.08 & 112.68 & -- & \textbf{32.14} & \textbf{3.51$\times$} \\
64 & 3,360 & 224.63 & 18.13 & 0.08 & 224.41 & -- & \textbf{62.20} & \textbf{3.61$\times$} \\
128 & 3,360 & 445.33 & 35.34 & 0.08 & 450.01 & -- & \textbf{121.99} & \textbf{3.69$\times$} \\

\bottomrule
\end{tabular}
\end{table*}

\begin{table*}[t]
\centering
\caption{Forward performance comparison on 32-H100.}
\label{tab:forward_only_32}
\begin{tabular}{c c c c c c c c c}
\toprule
B & Seq & $T_{comm}$ & $T_{comp}$ & CCR 
& Ring & TASP & CHARM+ & Speedup \\
\midrule

\multicolumn{9}{c}{\textbf{Seq = 126,976}} \\
\midrule
2 & 126,976 & 95.28 & 36.30 & 0.38 & 102.34 & 94.17 & \textbf{39.95} & \textbf{2.56$\times$} \\
4 & 126,976 & 186.17 & 70.03 & 0.38 & 208.12 & 184.59 & \textbf{77.02} & \textbf{2.70$\times$} \\
\midrule

\multicolumn{9}{c}{\textbf{Seq = 63,488}} \\
\midrule
2 & 63,488 & 49.95 & 12.50 & 0.25 & 46.32 & 48.54 & \textbf{19.80} & \textbf{2.34$\times$} \\
4 & 63,488 & 96.13 & 22.28 & 0.23 & 95.79 & 93.60 & \textbf{36.62} & \textbf{2.62$\times$} \\
8 & 63,488 & 187.15 & 41.96 & 0.22 & 193.04 & 184.12 & \textbf{70.62} & \textbf{2.73$\times$} \\
\midrule

\multicolumn{9}{c}{\textbf{Seq = 31,744}} \\
\midrule
2 & 31,744 & 26.70 & 6.28 & 0.24 & 22.59 & 25.38 & \textbf{11.47} & \textbf{1.97$\times$} \\
4 & 31,744 & 50.19 & 8.99 & 0.18 & 45.25 & 48.38 & \textbf{18.91} & \textbf{2.39$\times$} \\
8 & 31,744 & 96.07 & 15.58 & 0.16 & 92.39 & 93.35 & \textbf{35.27} & \textbf{2.62$\times$} \\
16 & 31,744 & 187.54 & 28.46 & 0.15 & 185.42 & 183.22 & \textbf{68.17} & \textbf{2.72$\times$} \\
\midrule

\multicolumn{9}{c}{\textbf{Seq = 15,872}} \\
\midrule
2 & 15,872 & 15.54 & 4.98 & 0.32 & 12.06 & 14.09 & \textbf{9.19} & \textbf{1.31$\times$} \\
4 & 15,872 & 26.73 & 5.09 & 0.19 & 22.29 & 25.24 & \textbf{10.93} & \textbf{2.04$\times$} \\
8 & 15,872 & 50.16 & 7.13 & 0.14 & 44.77 & 48.23 & \textbf{18.47} & \textbf{2.42$\times$} \\
16 & 15,872 & 96.20 & 11.93 & 0.12 & 89.27 & 93.30 & \textbf{34.64} & \textbf{2.58$\times$} \\
32 & 15,872 & 186.74 & 21.26 & 0.11 & 180.66 & 183.29 & \textbf{66.85} & \textbf{2.70$\times$} \\
\midrule

\multicolumn{9}{c}{\textbf{Seq = 7,936}} \\
\midrule
4 & 7,936 & 15.63 & 4.96 & 0.32 & 12.02 & 14.09 & \textbf{8.97} & \textbf{1.34$\times$} \\
8 & 7,936 & 27.24 & 5.01 & 0.18 & 22.31 & 25.31 & \textbf{10.83} & \textbf{2.06$\times$} \\
16 & 7,936 & 50.01 & 6.33 & 0.13 & 43.95 & 48.34 & \textbf{18.34} & \textbf{2.40$\times$} \\
32 & 7,936 & 96.10 & 10.17 & 0.11 & 89.66 & 93.28 & \textbf{34.44} & \textbf{2.60$\times$} \\
64 & 7,936 & 186.69 & 17.81 & 0.10 & 179.13 & 182.93 & \textbf{66.48} & \textbf{2.69$\times$} \\
\midrule

\multicolumn{9}{c}{\textbf{Seq = 3,968}} \\
\midrule
8 & 3,968 & 15.55 & 4.71 & 0.30 & 11.91 & 14.08 & \textbf{8.69} & \textbf{1.37$\times$} \\
16 & 3,968 & 27.24 & 5.12 & 0.19 & 22.29 & 25.23 & \textbf{10.90} & \textbf{2.04$\times$} \\
32 & 3,968 & 50.60 & 6.31 & 0.12 & 44.73 & 48.17 & \textbf{18.39} & \textbf{2.43$\times$} \\
64 & 3,968 & 96.26 & 9.56 & 0.10 & 89.71 & 93.11 & \textbf{34.44} & \textbf{2.60$\times$} \\
128 & 3,968 & 189.89 & 16.43 & 0.09 & 178.51 & 183.09 & \textbf{66.43} & \textbf{2.69$\times$} \\

\bottomrule
\end{tabular}
\end{table*}

\begin{table*}[t]
\centering
\caption{Backward performance comparison on 32-H100.}
\label{tab:backward_only_32}
\begin{tabular}{c c c c c c c c c}
\toprule
B & Seq & $T_{comm}$ & $T_{comp}$ & CCR 
& Ring & TASP & CHARM+ & Speedup \\
\midrule

\multicolumn{9}{c}{\textbf{Seq = 126,976}} \\
\midrule
2 & 126,976 & 259.62 & 102.57 & 0.40 & 280.48 & -- & \textbf{111.60} & \textbf{2.51$\times$} \\
4 & 126,976 & 521.26 & 196.56 & 0.38 & 559.74 & -- & \textbf{213.85} & \textbf{2.62$\times$} \\
\midrule

\multicolumn{9}{c}{\textbf{Seq = 63,488}} \\
\midrule
2 & 63,488 & 138.65 & 29.95 & 0.22 & 141.18 & -- & \textbf{46.30} & \textbf{3.05$\times$} \\
4 & 63,488 & 269.38 & 58.46 & 0.22 & 280.96 & -- & \textbf{86.70} & \textbf{3.24$\times$} \\
8 & 63,488 & 529.05 & 111.96 & 0.21 & 560.30 & -- & \textbf{166.44} & \textbf{3.37$\times$} \\
\midrule

\multicolumn{9}{c}{\textbf{Seq = 31,744}} \\
\midrule
2 & 31,744 & 70.47 & 10.16 & 0.14 & 70.57 & -- & \textbf{24.09} & \textbf{2.93$\times$} \\
4 & 31,744 & 140.16 & 19.39 & 0.14 & 141.71 & -- & \textbf{40.55} & \textbf{3.49$\times$} \\
8 & 31,744 & 275.78 & 37.57 & 0.14 & 280.92 & -- & \textbf{75.16} & \textbf{3.74$\times$} \\
16 & 31,744 & 536.42 & 73.45 & 0.14 & 559.34 & -- & \textbf{146.01} & \textbf{3.83$\times$} \\
\midrule

\multicolumn{9}{c}{\textbf{Seq = 15,872}} \\
\midrule
2 & 15,872 & 35.78 & 4.67 & 0.13 & 35.77 & -- & \textbf{14.92} & \textbf{2.40$\times$} \\
4 & 15,872 & 70.70 & 7.28 & 0.10 & 71.35 & -- & \textbf{22.11} & \textbf{3.23$\times$} \\
8 & 15,872 & 140.84 & 13.69 & 0.10 & 141.53 & -- & \textbf{37.75} & \textbf{3.75$\times$} \\
16 & 15,872 & 277.48 & 26.20 & 0.09 & 279.85 & -- & \textbf{70.62} & \textbf{3.96$\times$} \\
32 & 15,872 & 548.62 & 50.62 & 0.09 & 558.53 & -- & \textbf{137.84} & \textbf{4.05$\times$} \\
\midrule

\multicolumn{9}{c}{\textbf{Seq = 7,936}} \\
\midrule
4 & 7,936 & 35.82 & 3.71 & 0.10 & 35.18 & -- & \textbf{14.51} & \textbf{2.42$\times$} \\
8 & 7,936 & 70.72 & 5.82 & 0.08 & 71.06 & -- & \textbf{21.27} & \textbf{3.34$\times$} \\
16 & 7,936 & 140.74 & 10.90 & 0.08 & 141.24 & -- & \textbf{36.61} & \textbf{3.86$\times$} \\
32 & 7,936 & 278.53 & 20.68 & 0.07 & 278.38 & -- & \textbf{69.02} & \textbf{4.03$\times$} \\
64 & 7,936 & 550.51 & 39.76 & 0.07 & 556.25 & -- & \textbf{134.48} & \textbf{4.14$\times$} \\
\midrule

\multicolumn{9}{c}{\textbf{Seq = 3,968}} \\
\midrule
8 & 3,968 & 35.79 & 3.24 & 0.09 & 35.86 & -- & \textbf{13.97} & \textbf{2.57$\times$} \\
16 & 3,968 & 70.73 & 5.16 & 0.07 & 71.33 & -- & \textbf{20.95} & \textbf{3.41$\times$} \\
32 & 3,968 & 140.73 & 9.70 & 0.07 & 141.84 & -- & \textbf{36.35} & \textbf{3.90$\times$} \\
64 & 3,968 & 278.31 & 18.40 & 0.07 & 279.62 & -- & \textbf{68.40} & \textbf{4.09$\times$} \\
128 & 3,968 & 551.62 & 35.58 & 0.06 & 560.63 & -- & \textbf{133.05} & \textbf{4.21$\times$} \\
\bottomrule
\end{tabular}
\end{table*}





\end{document}